\documentclass[letterpaper]{article} 
\usepackage{aaai23}  
\usepackage{times}  
\usepackage{helvet}  
\usepackage{courier}  
\usepackage[hyphens]{url}  
\usepackage{graphicx} 
\usepackage{natbib}  
\usepackage{caption} 
\usepackage{soul}

\usepackage[figuresright]{rotating} 
\usepackage{microtype}
\usepackage{graphicx}
\usepackage{subfigure}
\usepackage{booktabs} 
\usepackage{wrapfig}
\usepackage{pifont}
\usepackage{multirow}
\usepackage[pagebackref=false, breaklinks=true, colorlinks, citecolor=citecolor, bookmarks=false]{hyperref}
\usepackage[table]{xcolor}

\definecolor{citecolor}{HTML}{0071BC}
\definecolor{linkcolor}{HTML}{ED1C24}

\usepackage{amsmath}
\usepackage{amssymb}
\usepackage{mathtools}
\usepackage{amsthm}
\usepackage{tcolorbox}

\DeclareMathOperator*{\argmax}{arg\,max}

\usepackage{bm}
\usepackage{amssymb}
\usepackage{subfigure}
\usepackage{mathtools}
\usepackage{amsthm}
\usepackage{color}
\usepackage{caption}
\usepackage{newfloat}
\usepackage{listings}
\usepackage{mleftright}
\usepackage{algorithm}
\usepackage{algorithmic}

\DeclareCaptionStyle{ruled}{labelfont=normalfont,labelsep=colon,strut=off} 
\floatstyle{ruled}
\newfloat{listing}{tb}{lst}{}
\floatname{listing}{Listing}
\usepackage{etoolbox}

\title{Lottery Pools: Winning More by Interpolating Tickets without Increasing Training or Inference Cost}
\author {
    Lu Yin\equalcontrib\textsuperscript{\rm 1},
    Shiwei Liu\equalcontrib\textsuperscript{\rm 1}\textsuperscript{\rm 2}\footnote{Corresponding author.},
    Meng Fang\textsuperscript{\rm 3},
    Tianjin Huang\textsuperscript{\rm 1},
    Vlado Menkovski\textsuperscript{\rm 1},
    Mykola Pechenizkiy\textsuperscript{\rm 1}}

\affiliations{
     \textsuperscript{\rm 1} Eindhoven University of Technology\\
     \textsuperscript{\rm 2} University of Texas at Austin\\
     \textsuperscript{\rm 3} University of Liverpool \\
   l.yin@tue,nl, s.liu3@tue.nl, meng.fang@liverpool.ac.uk, t.huang@tue.nl, v.menkovski@tue.nl, m.pechenizkiy@tue.nl

}
\begin{document}

\maketitle

\begin{abstract}
\looseness=-2 Lottery tickets (LTs) is able to discover accurate and sparse subnetworks that could be trained in isolation to match the performance of dense networks. Ensemble, in parallel, is one of the oldest time-proven tricks in machine learning to improve performance by combining the output of multiple independent models. However, the benefits of ensemble in the context of LTs will be diluted since ensemble does not directly lead to stronger sparse subnetworks, but leverages their predictions for a better decision. In this work, we first observe that directly averaging the weights of the adjacent learned subnetworks significantly boosts the performance of LTs. Encouraged by this observation, we further propose an alternative way to perform an ``ensemble'' over the subnetworks identified by iterative magnitude pruning via a simple interpolating strategy. We call our method \textbf{\textit{Lottery Pools}}. In contrast to the naive ensemble which brings no performance gains to each single subnetwork, Lottery Pools yields much stronger sparse subnetworks than the original LTs without requiring any extra training or inference cost. Across various modern architectures on CIFAR-10/100 and ImageNet, we show that our method achieves significant performance gains in both, in-distribution and out-of-distribution scenarios. Impressively, evaluated with VGG-16 and ResNet-18, the produced sparse subnetworks outperform the original LTs by up to \textbf{1.88\%} on CIFAR-100 and \textbf{2.36\%} on CIFAR-100-C; the resulting dense network surpasses the pre-trained dense-model up to \textbf{2.22\%} on CIFAR-100 and \textbf{2.38\%} on CIFAR-100-C. Our source code can be found at \url{https://github.com/luuyin/Lottery-pools.}





\end{abstract}

\section{Introduction}

Deep neural networks (DNNs) have revolutionized various machine learning fields with expressive performance~\citep{lecun1989handwritten,Alexnet,simonyan2014very,resnet,silver2016mastering,dosovitskiy2020image,brown2020language,radford2021learning,fedus2021switch,jumper2021highly}. While achieving increasingly compelling results, a large concern is the massive parameter count that in billions, even trillions resulting heavy burden on environmental and financial systems~\citep{garcia2019estimation,schwartz2020green,patterson2021carbon,zhang2022survey}. That motivates many techniques toward the efficiency of DNNs. Among them, sparsity is a leading approach that largely preserves the model performance while achieving appealing compression rates~\citep{mozer1989using,han2015deep,molchanov2016pruning,liu2022unreasonable,chen2022coarsening,sun2021load,yuan2021mest}. A recent work on the Lottery Tickets (LTs)~\citep{frankle2018lottery} discovers the existence of sparse subnetworks within a standard network which can be trained in isolation to match the accuracy of the dense counterpart. These lottery tickets are empirically obtained by iterative magnitude pruning (IMP) at the random dense initialization or early training points called ``rewinding''~\citep{frankle2020linear}. Since being proposed, LTs has become the leading approach to reducing model size while preserving accuracy.

However, LTs is a rather costly process involving multiple iterations of pruning-and-retraining, and once the subnetworks at the target sparsity are reached, the previous subnetworks with lower sparsity are commonly discarded, leading to a big waste of computation. One of the traditional ways that can directly benefit from multiple models in machine learning is ensemble. Ensemble~\citep{hansen1990neural,levin1990statistical,fort2019deep} is well-known for its compelling performance improvements over independently trained, single networks by combining the predictions of the latter. Yet, ensemble does not directly lead to stronger sparse subnetworks but leverages their predictions to make a better decision. Hence, the benefits of ensemble in the context of LTs will be diluted.


In this paper, we build an efficient and accurate alternative to the naive LTs ensemble that yields subnetworks outperforming the original LTs by a large margin in both, in-distribution and out-of-distribution scenarios. We first observe that directly interpolating weights of the adjacent LTs subnetworks improves the performance of LTs. Inspired by this observation, we sequentially interpolate the weights of the natural ``byproducts'' of LTs (i.e., the previous subnetworks identified by IMP) with the target subnetwork if they improve accuracy on held-out data, following the recent emerging weight averaging techniques~\citep{izmailov2018averaging,wortsman2022model,rame2022diverse}. This simple interpolating step is able to produce much stronger sparse and dense networks, without incurring any additional training and inference costs. We call our approach ``Lottery Pools''\footnote{ Lottery pools refers to a group of people who purchase lottery tickets together to get better odds of winning a lottery. We borrow this concept to highlight that we combine (interpolate) multiple LTs subnetworks into a stronger one with higher accuracy.}. Unlike the original LTs, Lottery Pools harnesses the advantage of all the LTs subnetworks, in turn, to further boost the performance of each of them. This property significantly increases the utility of the identified subnetworks compared with the original LTs, where the previous well-learned subnetworks are usually discarded. Overall, our contributions are summarised as follows: 




\begin{itemize}

    \item  \textbf{Simple weight interpolation between two adjacent subnetworks boosts the performance of LTs.} We surprisingly find that adjacent subnetworks of LTs can be linearly interpolated or even averaged into a single subnetwork with higher accuracy while maintaining the same sparsity (shown in Figure~\ref{fig:ACC_heatmap}). 

    \item  \textbf{Towards stronger LTs subnetworks.} Encouraged by the above observation, we propose Lottery Pools that selectively interpolate multiple subnetworks into a single subnetwork. Lottery Pools is able to construct stronger subnetworks with much higher accuracy, while \underline{maintaining the original sparsity level}. Simple as it is, we show that Lottery Pools (Interpolation) surpasses the original LTs by 1.88\% and 1.72\%  on CIFAR-100 with VGG-16 and ResNet-18 respectively.
    
    \item \textbf{Towards stronger dense networks.} Besides the improved sparse subnetworks, we can also construct a stronger dense network by averaging the LTs tickets back to the pre-trained dense networks. Our reinforced dense network outperforms the original dense ResNet-18 by 2.22\% and the original dense VGG-16 by 1.69\% on CIFAR-100.
    
    \item \textbf{Towards Stronger in-distribution and out-of-distribution performance.} Thanks to the ``ensemble'' property of Lottery Pools, the enhanced (sub)networks enjoy a remarkable performance gain over the original LTs/dense model in both in-distribution (ID) predictive accuracy and out-of-distribution (OoD) robustness.
    

\end{itemize}

\section{Related Work}

\textbf{Lottery ticket hypothesis.} 
Lottery ticket (LTs)~\citep{frankle2018lottery} conjectures that
there exist sparse subnetworks called winning tickets within
a dense network, whose performance can match with the
dense network when training from the same initialization. 
Later, weight/learning rate rewinding techniques~\citep{frankle2020linear,renda2020comparing} was proposed to 
scale up LTs to larger networks and datasets.  \citet{evci2022gradient} demonstrates that training LTs solutions with the same initialization converge to the same basin as the original pruning method that they are derived from. LTs has inspired many follow-up works to understand and extend LTs.~\citet{zhou2019deconstructing,ramanujan2020s} successfully found winning tickets at the initialization even without training.~\citet{morcos2019one} unveiled that the winning tickets discovered using larger datasets consistently transferred better than those generated using smaller datasets. Besides the original ImageNet classification~\citep{frankle2018lottery}, the existence of winning tickets has been broadly verified under diverse fields, such as natural language processing~\citep{gale2019state,chen2020lottery}, generative adversarial networks~\citep{chen2021gans}, and reinforcement learning~\citep{yu2019playing}.   Unlike LTs, Lottery Pools takes advantage of all the sparse subnetworks to construct stronger subnetworks without increasing any extra training or inference time.

\looseness=-1 \textbf{Weight averaging.} Model weight averaging has been widely studied in convex optimization and neural networks~\citep{ruppert1988efficient,polyak1992acceleration,zhang2019lookahead}. Stochastic Weight Averaging (SWA)~\citep{izmailov2018averaging} and Exponential Moving Average (EMA)~\citep{polyak1992acceleration} average checkpoints along a single optimization trajectory and can roughly match the prediction ensemble performance.~\citet{yin2022superposing} further generated SWA in the context of sparse training without any pretraining steps. Greedy soup~\citep{wortsman2022model} averages independent dense models across different runs, providing notable improvements. 

\looseness=-1 Weight interpolation, as a more general case of weight averaging, draws explosive interest from the community recently.~\citet{nagarajan2019uniform} empirically observed that there exists a linear path between the solutions learned on MNIST dataset with the same initialization.~\citet{neyshabur2020being} shown that two models fine-tuned from the same pre-trained model can be linearly connected to match the performance of the single model.~\citet{wortsman2022model} proposed learned soup recipe that learns model interpolation by AdamW~\citep{loshchilov2017decoupled}. Weight interpolation has also been adopted to improve the accuracy of the patching task without compromising accuracy on the supported tasks and transfer learning~\citep{patching}.~\citet{frankle2020linear} introduced linear mode connectivity to study the instability of neural networks to the SGD noise introduced during training. They demonstrated that sparse subnetworks discovered by LTs can match the performance of the dense network only when they are stable to SGD noise. Following~\citet{frankle2020linear}, we discover that two linearly connected subnetworks can be  interpolated, leading to a more accurate subnetwork without any extra costs. 

\textbf{Ensemble.} Ensembles~\citep{hansen1990neural,levin1990statistical} of neural networks have received large success in terms of the in-distribution accuracy~\citep{perrone1992networks,breiman1996bagging,Dietterich2000EnsembleMI}, uncertainty estimation~\citep{lakshminarayanan2017simple,wen2020batchensemble}, and out-of-distribution robustness~\citep{Ovadia2019CanYT,gustafsson2020evaluating}. Very recently,~\citet{liu2021deep} proposed an efficient ensemble framework that combines the
predictions of multiple individual subnetworks, surpassing the generalization performance of the naive ensemble. Nevertheless, the ensemble requires performing a forward pass for each model, leading to extra costs.

  \begin{table}
    \centering
 
            \caption{The primary methods contrasted in this work.
            ${\bm{\mathrm{\tilde\theta}}}$ is a subnetwork learned by IMP from different iteration. Cost refers to the memory and computes requirements during inference
            relative to a single model. 
            All methods require the same training.}
      
                \resizebox{0.35\textwidth}{!}{
                \begin{tabular}{lcc} 
                \toprule
                 Method & Formulation & Cost \\
                 \midrule
                 Lottery Tickets & $f(\bm{x}; {\bm{\mathrm{\tilde\theta}}})$  & $\mathcal{O}(1)$  \\
                 Naive Ensemble & $\frac{1}{k}\sum_{i=1}^k f(\bm{x}; {\bm{\mathrm{\tilde\theta}}}_{k}) $ & $\mathcal{O}(k)$ \\ %
                 Lottery Pools & Algorithm~\ref{alg:Aeraging_recipe} & $\mathcal{O}(1)$  \\ 

                 \bottomrule
                \end{tabular}}
            
            \label{tab:methods}
        \end{table}

\section{Methodology}
\label{sec:metho}

In this section, we introduce ``Lottery Pools'', a simple weight interpolation approach that constructs more accurate subnetworks than LTs, without requiring any extra training or inference cost.  Different from previous works on interpolation, our goal is to boost the performance of LTs subnetworks (including the pre-trained dense network) while maintaining their desirable sparsity.

We first recap the concept of the lottery ticket hypothesis. Then, we show that directly averaging LTs subnetworks identified at two adjacent IMP iterations leads to stronger subnetworks. Finally, we introduce our Lottery Pools, which is able to improve the accuracy of the original LTs by interpolating subnetworks obtained across different IMP iterations.

\textbf{Recapping the lottery ticket hypothesis.} The lottery ticket hypothesis~\citep{frankle2018lottery} indicates that there exist subnetworks (winning tickets) within dense network initialization such that those lottery tickets could be trained in isolation to the matching performance of their dense counterparts.  

To be specific, we denote neural network with parameters $\bm{\mathrm{\theta}}\in \mathbb{R}^d$ as $f(\bm{x}; \bm{\mathrm{\theta}})$,  after $j$  steps of training, there exists a sparse subnetwork characterized by the binary mask $m$ such that $f(\bm{x}; m\odot \bm{\mathrm{\theta}}_j)$  will perform as well as $f(\bm{x}; \bm{\mathrm{\theta}})$ after training. The initial results show LTs hold when the subnetwork is trained with the original initialization, i.e. $j=0$.  Later results~\cite{frankle2020linear} state a rewinding step to a pre-train point ($j>0$) is required on larger datasets. The overall training procedure of LTs is given in Algorithm~\ref{alg:IMP}. 

\begin{algorithm}[tb]
 
   \caption{Pseudocode of LTs}
   \label{alg:IMP}
\begin{algorithmic}[1]
\REQUIRE   Randomly initialization network  $\bm{\mathrm{\theta}}\in \mathbb{R}^d$, binary mask $m$, IMP iterations $T$, training steps used for rewinding $j$, pruning rate $p$.

    \STATE Train $\bm{\mathrm{\theta}}$ to completion; save the weights  at $j$ steps $\bm{\mathrm{\theta}}_j$.
    \FOR{pruning iteration $t \in \{1, \ldots, T\}$} 
        
    \STATE Prune the lowest magnitude $p$ of weights and update mask $m^t$.
    
    \STATE Train  $m^t \odot \bm{\mathrm{\theta}}_j$ to completion.

    \ENDFOR

\end{algorithmic}
\end{algorithm}

    


\begin{figure}[htbp]
\centering
    \subfigure[ResNet-18 w/ Rewinding]{
        \includegraphics[width=0.22\textwidth]{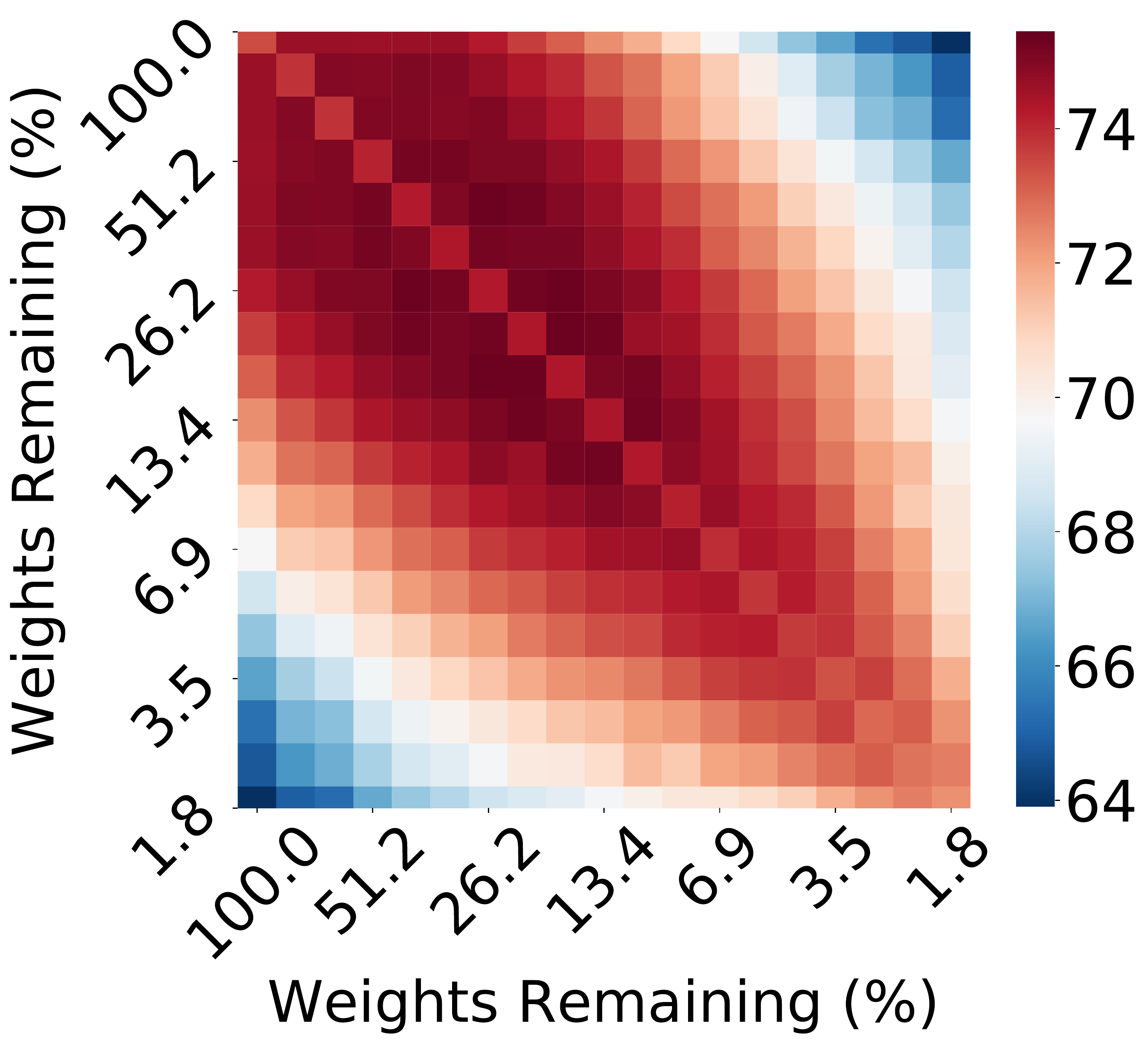}
    }
     \subfigure[ResNet-18 w/o Rewinding]{
        \includegraphics[width=0.22\textwidth]{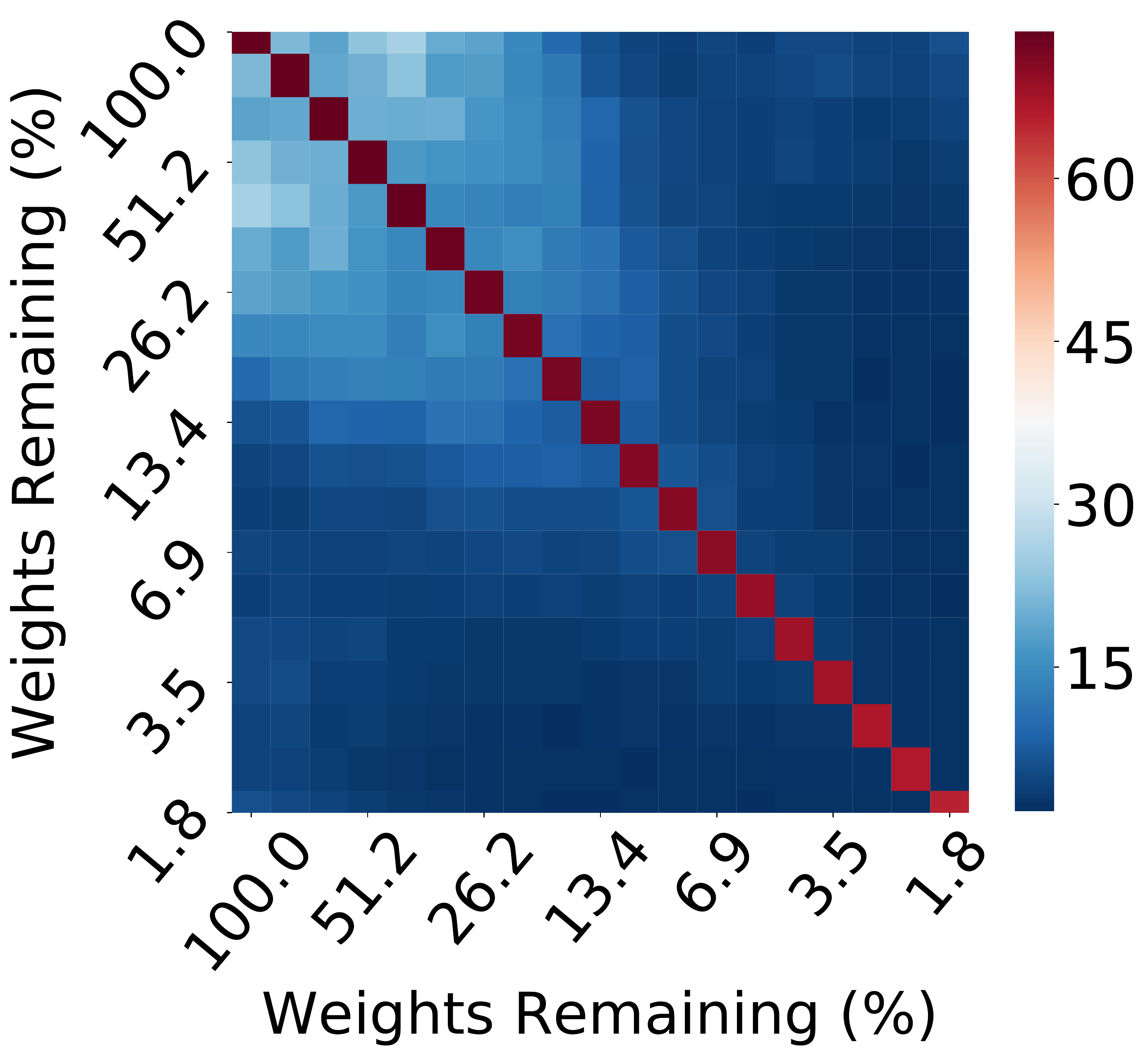}
    }
 
\caption{Accuracy heat map of the averaged LTs on CIFAR-100. Each cell refers to the test accuracy of the averaged subnetworks using the LTs under the sparsity of X-axis and Y-axis. If the averaged subnetwork decreases in sparsity, we prune it to the higher sparsity of its parents.}
    \label{fig:ACC_heatmap}

\end{figure}

\begin{algorithm*}[tb]
  \caption{Lottery Pools Interpolation Recipe}
  \label{alg:Aeraging_recipe}
   
\textbf{Input}: The original learned sparse subnetwork ${\bm{\mathrm{\tilde\theta}}}_{t}$ from LTs,  Candidate Lottery Pools $ \mathcal{S}_t = \{  {\bm{\mathrm{\tilde\theta}}_{t\text{-}1}},{\bm{\mathrm{\tilde\theta}}_{t\text{+}1}},{\bm{\mathrm{\tilde\theta}}_{t\text{+}2}},\cdots \}$,  Candidate Coefficient Pools $ \mathcal{S}_c = \{ {\mathrm{\alpha_1 }}, \ldots,{\mathrm{\alpha_n }}  \}$, The interpolated subnetwork ${\bm{\mathrm{\tilde\theta}}_{Inter}}$ during greedy search. 

\textbf{Output}: Final interpolated subnetwork ${\bm{\mathrm{\tilde\theta}}_{best}}$ has the same sparsity with ${\bm{\mathrm{\tilde\theta}}_t}$.
\begin{algorithmic}[1]

    \STATE $ \mathcal{S}_t  \gets \{  {\bm{\mathrm{\tilde\theta}}_{t\text{-}1}},{\bm{\mathrm{\tilde\theta}}_{t\text{+}1}},{\bm{\mathrm{\tilde\theta}}_{t\text{+}2}},\cdots \}$  \COMMENT{ Create Candidate Lottery Pools, and sort all candidates by their adjacence to  ${\bm{\mathrm{\tilde\theta}}_t}$ }
    \STATE  { $ \mathcal{S}_c  \gets  \{ {\mathrm{\alpha_1 }}, \ldots,{\mathrm{\alpha_n }}  \}$ }
    \COMMENT {Create Candidate Coefficient Pools }

    \STATE ${\bm{\mathrm{\tilde\theta }}_{best}}   \gets  {\bm{\mathrm{\tilde\theta}}}_t$

    \STATE  {\bfseries for} ${\bm{\mathrm{\tilde\theta}}_i}  \in \mathcal{S}_t  $  {\bfseries do}   \COMMENT{ Greedily search the candidate LTs subnetworks for interpolation  }

    \STATE \ \ \ $ {\alpha}_{best} \gets   \argmax_{j} \mathsf{ValAcc}\mleft(    \mathsf{MagnitudePruning} \mleft( \mathrm{\alpha}_j\bm{\mathrm{\tilde\theta}}_{best} + (1-{\mathrm{\alpha}_j})  \bm{\mathrm{\tilde\theta}}_{i}  \mright)\mright), {\mathrm{\alpha}_j}  \in \mathcal{S}_c$     
          \COMMENT{Search for the best coefficient}

    \STATE \ \ \  ${\bm{\mathrm{\tilde\theta}}_{Inter}} \gets \mathsf{MagnitudePruning} \mleft( \mathrm{\alpha}_{best}\bm{\mathrm{\tilde\theta}}_{best} + (1-{\mathrm{\alpha}}_{best})  \bm{\mathrm{\tilde\theta}}_{i} \mright) $  \COMMENT{Interpolating using  ${\alpha}_{best}$ }

    \STATE  \ \ \ {\bfseries if} $ \mathsf{ValAcc}\mleft({\bm{\mathrm{\tilde\theta}}_{Inter}} \mright)\geq \mathsf{ValAcc} \mleft( {\bm{\mathrm{\tilde\theta}}_{best}}  \mright) $ 
    
    \STATE \ \ \ \ \ \  \ \ \   {\bfseries then}     ${\bm{\mathrm{\tilde\theta}}_{best}} \gets    {\bm{\mathrm{\tilde\theta}}_{Inter}}$ 
    \COMMENT{Update the best interpolated subnetwork}
    

    \STATE  {\bfseries end for}

\end{algorithmic}
\end{algorithm*}

\textbf{Simple weight averaging boosts the performance of LTs.} Recently, the work on model soup~\citep{wortsman2022model} shows that averaging  models fine-tuned from the same pre-trained model provides substantial performance improvements. Since the LTs subnetworks at different sparsities are also fine-tuned from the same pre-trained dense model, we hypothesize that these subnetworks can also be averaged for better performance. 

Let's simplify the sparse subnetwork parameters under a mask  $m^t \odot {\bm{\mathrm{\theta}}}$ as  $\tilde{\bm{\mathrm{\theta}}}$ for simplicity. To verify our hypothesis, we average two LTs subnetworks with different sparsities (assuming sparsity of $\tilde{\bm{\mathrm{\theta}}}_1$ is higher than sparsity of  $\tilde{\bm{\mathrm{\theta}}}_2$) across all the learned subnetworks, i.e., $\frac{\tilde{\bm{\mathrm{\theta}}}_1 + \tilde{\bm{\mathrm{\theta}}}_2}{2}$. To address the sparsity decrease caused by average, we further use magnitude pruning $\frac{\tilde{\bm{\mathrm{\theta}}}_1 + \tilde{\bm{\mathrm{\theta}}}_2}{2}$ to the same sparsity as $\tilde{\bm{\mathrm{\theta}}}_1$ following~\citep{yin2022superposing}. We consider two widely used settings for LTs: with rewinding and without rewinding. We report the results of ResNet-18 on CIFAR-100 within a heat map in Figure~\ref{fig:ACC_heatmap} and put the results of VGG-16 in the Appendix. 

As we can see, \textit{rewinding matters for the improved performance of weight averaging}. With rewinding, simple averaging subnetworks from nearby IMP iterations could achieve better performance than the original LTs subnetworks (the diagonal cells), and the closer two subnetworks are located, the larger performance gains the averaged subnetworks tend to achieve.  In stark contrast,  we observe a significant accuracy drop without rewinding, across all sparsities. This observation is in line with the findings from~\citet{frankle2020linear} that large-scale settings are unstable to SGD noise at initialization according to linear interpolation. Thereby, we confirmed that our hypothesis holds in the context of rewinding.

\begin{figure}[ht]
\centering
    \subfigure{
        \includegraphics[width=0.5\textwidth]{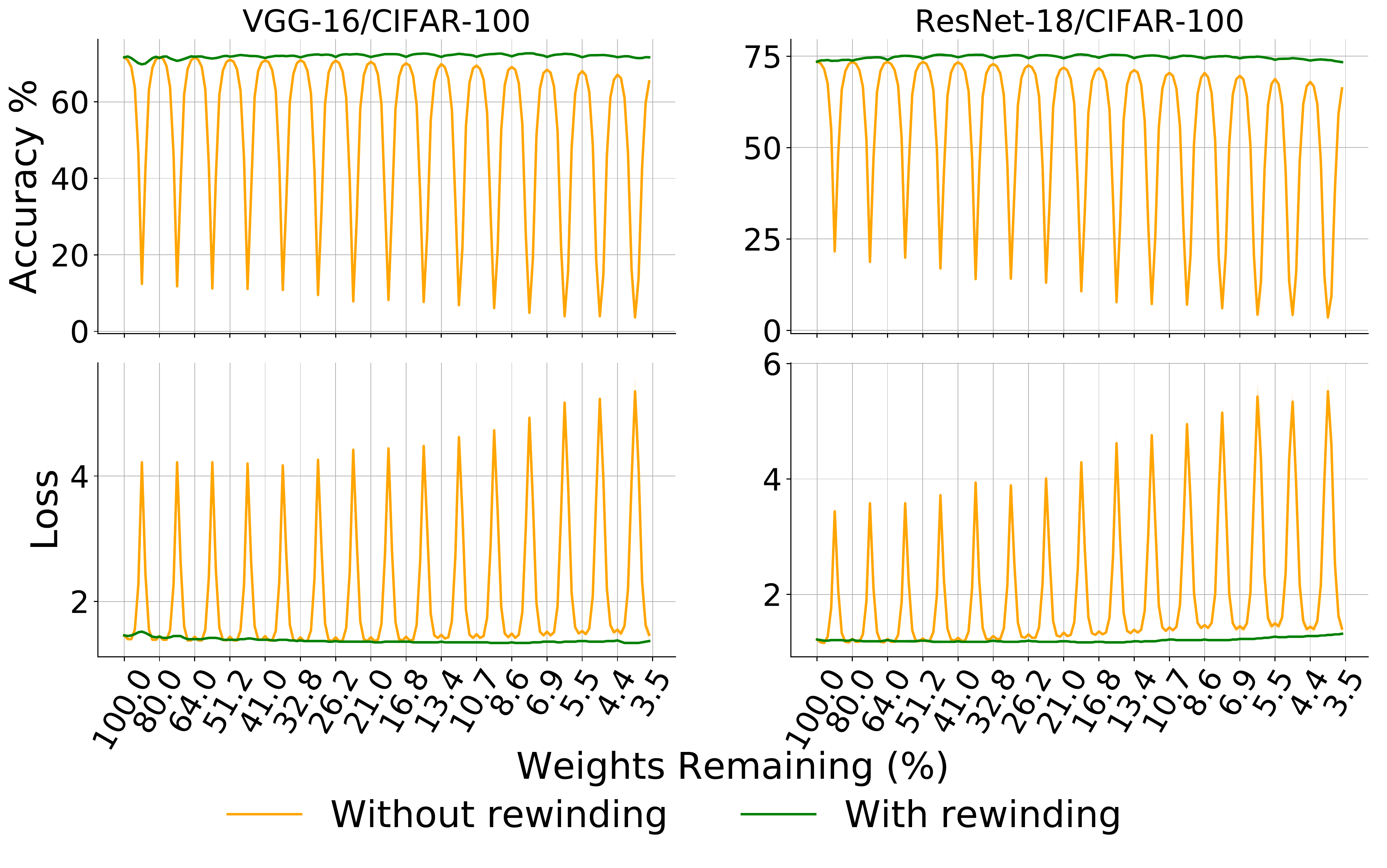}
    }
    
\caption{Liner interpolation between  learned subnetworks  from dense network to extremely low density.}
    \label{fig:Liner_interpolation}

\end{figure}

\subsection{Lottery Pools}
Inspired by the above observation, we introduce Lottery Pools, a simple weight interpolation approach on LTs that leverages the subnetworks obtained across different IMP iterations. Lottery Pools has two key ideas: (1) Interpolating weights instead of simply averaging; (2) Sequentially searching over all the candidate LTs subnetworks and coefficients for interpolation. 

Firstly, Lottery Pools goes beyond weight averaging and probes a more general variant of weight connection -- weight interpolation.  Linear weight interpolation has been previously used to study dense networks~\citep{nagarajan2019uniform,neyshabur2020being}. Here, we adopt it to improve the accuracy of sparse LTs subnetworks.
We follow~\cite{frankle2020linear} and determine that two LTs subnetworks are linear mode connected if there exists a linear path between them. The two subnetworks thus could be interpolated for accuracy improvement. The subnetwork created by interpolating is given by:

 \begin{equation}
\bm{\mathrm{\tilde\theta}}_{inter} =  \mathrm{\alpha }\bm{\mathrm{\tilde\theta}}_{1} + (1-{\mathrm{\alpha }})  \bm{\mathrm{\tilde\theta}}_{2}
\label{eq:interpolation}
\end{equation}
We argue that directly averaging two subnetworks ($\alpha=0.5$) might not be the optimal option, since the local linearly-connected minimum may lie in the sides of the linear path rather than in the right middle. To prove this, we linearly interpolate two adjacent LTs subnetworks, with various coefficients $\alpha \in [0,1]$. We set the increments of $\alpha$ as $0.1$ to create 9 interpolated subnetworks between two LTs subnetworks. The test loss/error are reported in Figure~\ref{fig:Liner_interpolation}. The best accuracy is achieved at the middle points of the linear path in most cases (the middle part of the green lines) with exceptions at the highest and lowest sparsities (the endpoints of the green lines) where weight average achieves no better accuracy than interpolation. Again, interpolation without weight rewinding (yellow lines) fails to find such linear paths between two subnetworks.


Moreover, Figure~\ref{fig:ACC_heatmap} and~\ref{fig:Liner_interpolation} present that the accuracy of the interpolated subnetworks varies across different subnetworks pairs and different interpolation coefficients. While it is possible to adopt gradient-based optimization to learn the optimal subnetworks and coefficients, the cost is rather expensive. We instead choose a more practical way: greedily searching for interpolated subnetworks and the corresponding coefficients. Given a target LTs subnetwork  ${\bm{\mathrm{\tilde\theta}}_{t}}$ learned at the $t$ iteration of IMP, we sequentially interpolate it with the rest of the subnetworks (Candidate Lottery Pools). For each candidate subnetwork, we search the best coefficients $\alpha$ over 11 candidates from 0.05 to 0.95 (Candidate Coefficient Pools). Please refer to the Appendix for more details. We only keep the incoming subnetwork for interpolation if its accuracy on the held-out set does not decrease. 

Instead of loading all the candidate subnetworks in the memory~\cite {wortsman2022model}, we apply a more memory-friendly search approach. Specifically, we iteratively interpolate one of the subnetworks in the Candidate Lottery Pools with our target subnetwork ${\bm{\mathrm{\tilde\theta}}_{t}}$ and let the resulting subnetwork being our new target subnetwork, i.e. ${\bm{\mathrm{\tilde\theta}}}_{t} \leftarrow {\bm{\mathrm{\tilde\theta}}_{inter}}$, until we have searched all the subnetworks.  
This operation allows us to accomplish the interpolation operation across all the candidate subnetworks by maintaining only one extra copy of the interpolated weights.

In summary, Lottery Pools is a two-step procedure for constructing stronger LTs subnetworks. \textbf{Step 1:} Perform the standard Lottery Tickets method; \textbf{Step 2:} Linearly interpolate the original LTs subnetworks to produce stronger sparse subnetworks. Please note that we adopt magnitude weight pruning to remove the weights with the smallest magnitude after each interpolation to maintain the same sparsity level as the original LTs.

\section{Experiments}





\begin{figure*}[ht]
\centering
    \subfigure[Test accuracy {\%} of the original LTs and Lottery Pools on CIFAR-10/100.]{
        \includegraphics[width=0.99\textwidth]{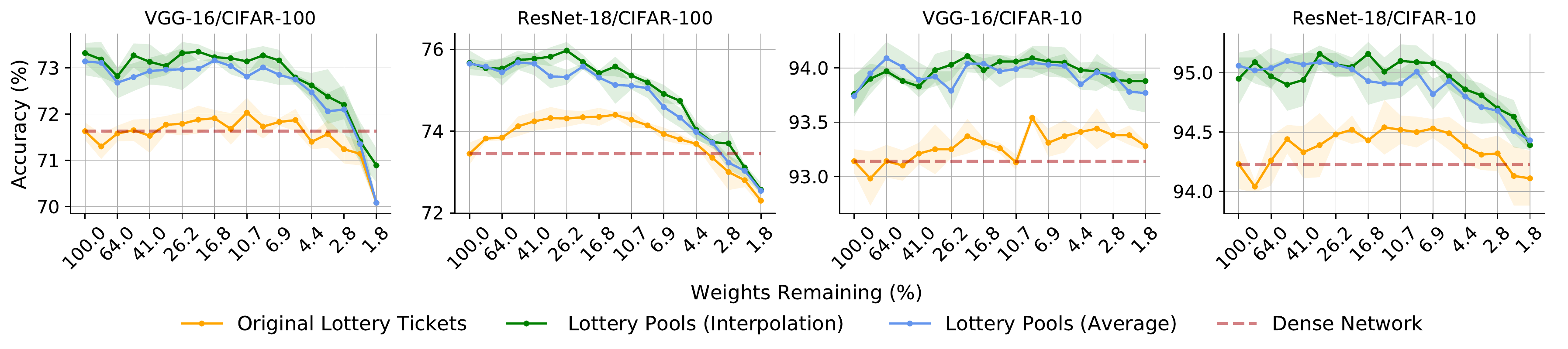}
        \label{fig:Cifar_acc}
    }
  
    \subfigure[Comparison with the strong weight averaging baselines: SWA and EMA.]{
        \includegraphics[width=0.99\textwidth]{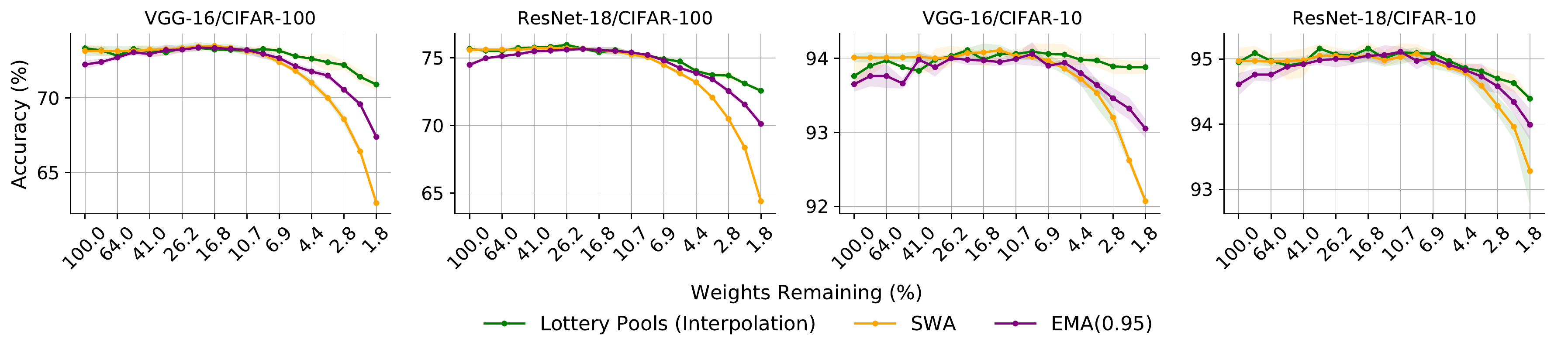}
        \label{fig:baseline}
    }    
    
    \subfigure[Test accuracy {\%} of the original LTs and Lottery Pools on CIFAR-10-C and CIFAR-100-C.]{
        \includegraphics[width=0.99\textwidth]{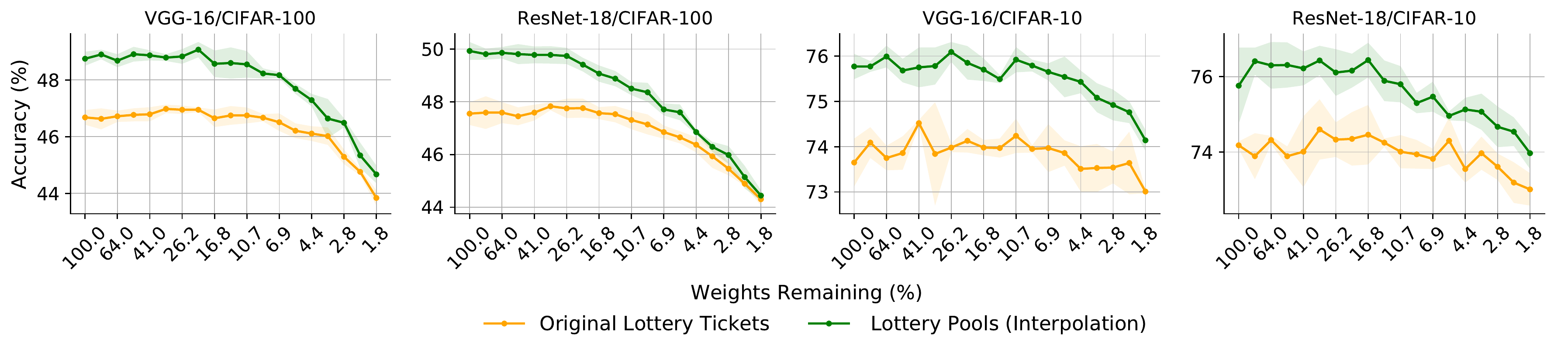}
        \label{fig:ood_acc}
    }

\caption{Evaluation of Lottery Pools.}

\end{figure*}

To verify the effectiveness of Lottery Pools, we evaluate it with three popular model structures VGG-16, ResNet-18 and ResNet-34 on various datasets, including CIFAR-10 and CIFAR-100 and ImageNet.

\begin{table}[h]
    \centering
    
    \caption{Implementation details, including:  IMP iteration count, rewinding epochs, learning rate (LR), batch size (BS),  learning rate drop (LR Drop), training epochs (Epoch), etc.}
    \label{tab:setting}
    \resizebox{0.48\textwidth}{!}{
    \begin{tabular}{cc|c|c|c|c|c|c|c}
        \toprule
         Network  & Dataset & Epoch & BS & LR & LR Drop, Epochs  & Warmup &  Rewinding  &  IMP iterations \\ \midrule
         ResNet-18& CIFAR-10/100  &182  & 128 & 0.1 & 10x, [91, 136]  & - & 9 Epochs & 19   \\\midrule
         VGG-16& CIFAR-10/100   &182& 128 & 0.1 &  10x, [91, 136]  & - &  9 Epochs &  19  \\\midrule
          ResNet-18/34& ImageNet  &90 & 1024 & $  0.4 $ &  10x, [30,60,80]   & 5 Epochs & 5 Epochs & 9   \\
         
         \bottomrule
    \end{tabular}}
\end{table}
\textbf{Experiments setup.} 
Since our method directly performs weight interpolation over the subnetworks produced by LTs rewinding, our most direct baseline is the standard LTs rewinding. Following the common rewinding setting used in~\citet{frankle2020linear,chen2020lottery}, we rewind the LTs roughly to the 5\% training time. We summarize the  implementation details for LTs in table~\ref{tab:setting}. 
To highlight the performance difference between weight interpolation and weight averaging, we implement two variants of Lottery Pools: Lottery Pools (Interpolation) and Lottery Pools (Average), the latter applies direct averaging instead of interpolation in the second step of Lottery Pools. The results are illustrated in Figure~\ref{fig:Cifar_acc}. All the reported results are averaged over 3 independent runs.


\begin{figure*}[ht]
\centering
\includegraphics[width=0.99\textwidth]{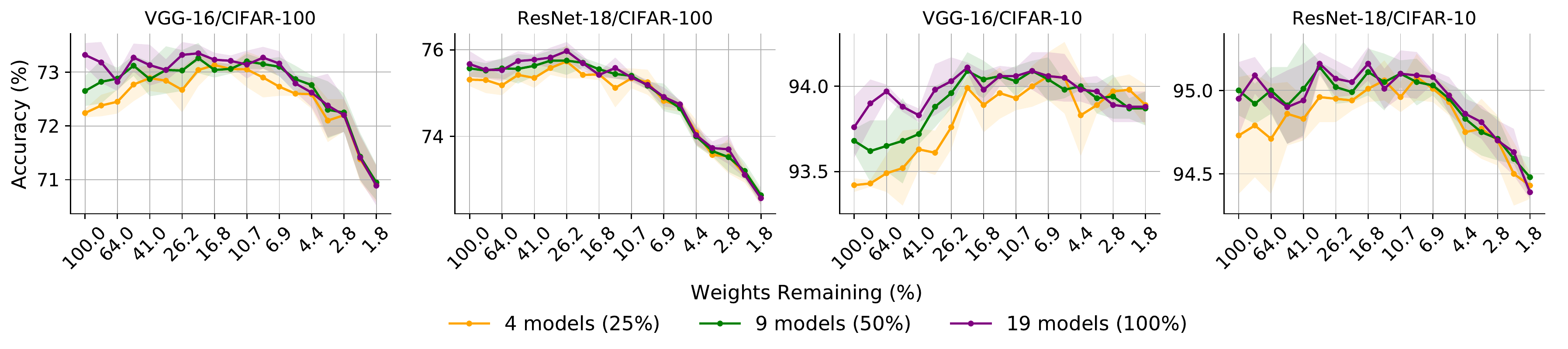} 

\caption{Test accuracy {\%} of  different candidate model count.}

\label{fig:model_count}

\end{figure*}

\begin{figure*}[ht]
\centering
\includegraphics[width=0.99\textwidth]{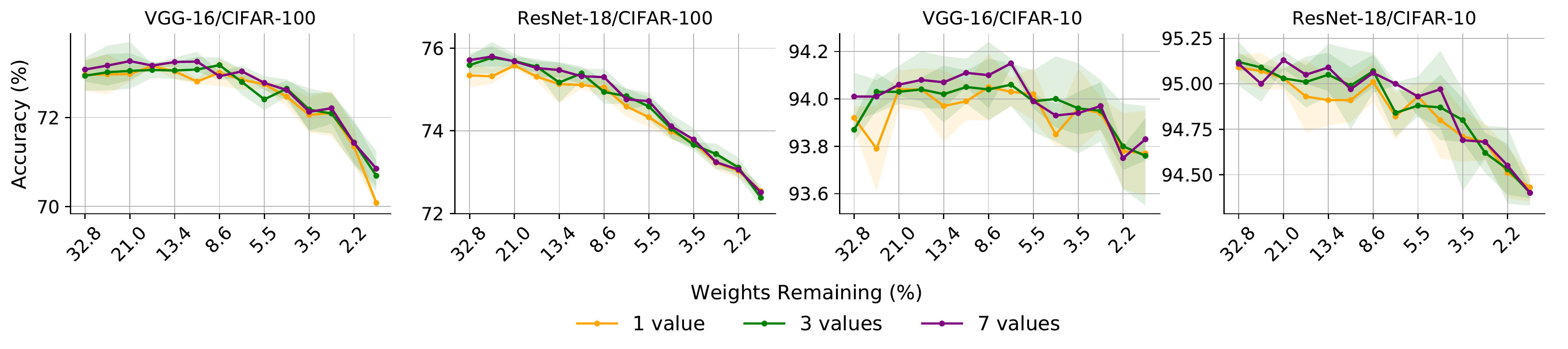} 

\caption{Test accuracy {\%} of  different candidate interpolation coefficient count.}

\label{fig:value_count}
\end{figure*}

\textbf{Comparison with the original LTs.} Overall, we see a clear performance gain from Lottery Pools over the original  LTs under different sparsity levels (including the dense network), and Lottery Pools (Interpolation) achieves better performance than Lottery Pools (Average) due to the searched optimal interpolation value.
Impressively, Lottery Pools (Interpolation) achieves up to 1.88\% and 1.72\% accuracy increase over the original LTs with VGG-16 and ResNet-18 on CIFAR-100, respectively. Even on the relatively more saturated CIFAR-10, we still observe up to 0.93\% and 1.05\% performance gains with VGG-16 and ResNet-18. We highlight that Lottery Pools also outperforms LTs in even extremely sparse situations. For instance, Lottery Pools brings 0.81\% higher accuracy over the LTs with VGG-16 on CIFAR-100 and 0.6\%  higher accuracy on CIFAR-10, with only 1.8\% weights. It is quite encouraging to see that Lottery Pools can still improve performance when the interpolating space is extremely small.   Besides, by averaging the learned LTs subnetworks back to the pre-trained dense model, Lottery Pools (interpolation) could also construct stronger dense networks, which outperform the original dense ResNet-18 by 2.22\%  and the original dense VGG-16 by 1.69\% on CIFAR-100. We report the results with ResNet-18/34 on ImageNet in the Appendix due to the limited space, where Lottery Pools also consistently outperforms LTs.

\textbf{Comparison with weight averaging baselines.} We further compare our method with two strong weight averaging baselines: SWA and EMA. SWA~\citep{izmailov2018averaging} averages the weights of multiple networks along a single optimization trajectory. By setting the averaging coefficient to $\frac{1}{n+1}$ where $n$ is the current model number, it achieves the same results of averaging across all the models while maintaining the memory consumption as just two DNNs. The exponential moving average (EMA)~\cite{polyak1992acceleration,kingma2014adam,karras2017progressive}  average the weights of a series of models exponentially using a fixed decay factor, which was set to 0.95 in this experiment. As these baselines are all designed for dense model averaging, the same as Lottery Pools, we alter EMA and SWA by pruning the interpolated model to the sparsity of the original LHs during interpolating.    In fact, both SWA and EMA could be regarded as special cases of linear interpolation that use specific averaging coefficients and decay factors for weight interpolating.

As shown in Figure~\ref{fig:baseline}, our method clearly outperforms the other baselines by a large margin, especially in highly sparse situations. That apparently comes from two possible reasons. Firstly, EMA and SWA use pre-chosen coefficient values, which might not generate all the subnetworks, whereas Lottery Pools searches for the optimal values for each subnetwork. Secondly, Lottery Pools greedily searches for potential subnetworks to interpolate, which can eliminate the negative effects of subnetworks that are adding networks within the different loss basins for interpolation. 

\textbf{Out-of-distribution robustness.} Beyond the in-distribution accuracy, we highlight that Lottery Pools also improves the performance of LTs in the OoD scenario. We train Lottery Pools with standard CIFAR-10 and CIFAR-100 and test it on CIFAR-10-C and CIFAR-100-C, respectively. As shown in Figure~\ref{fig:ood_acc}, Lottery Pools improves the OoD robustness than the original LTs by a large margin with both dense and sparse subnetworks. Remarkably, it improves the dense ResNet-18 by 2.38\% on CIFAR-100, and by 2.12\% with VGG-16 on CIFRA-10. For sparse subnetworks, Lottery Pools achieves up to 2.27\%  performance gain with VGG-16, and 2.36\% with ResNet-18 on CIFAR-100. This result indicates that our interpolated subnetworks are able to inherit the appealing properties of model ensemble, e.g., good OoD robustness.

\begin{figure*}[ht]
\centering
\includegraphics[width=0.99\textwidth]{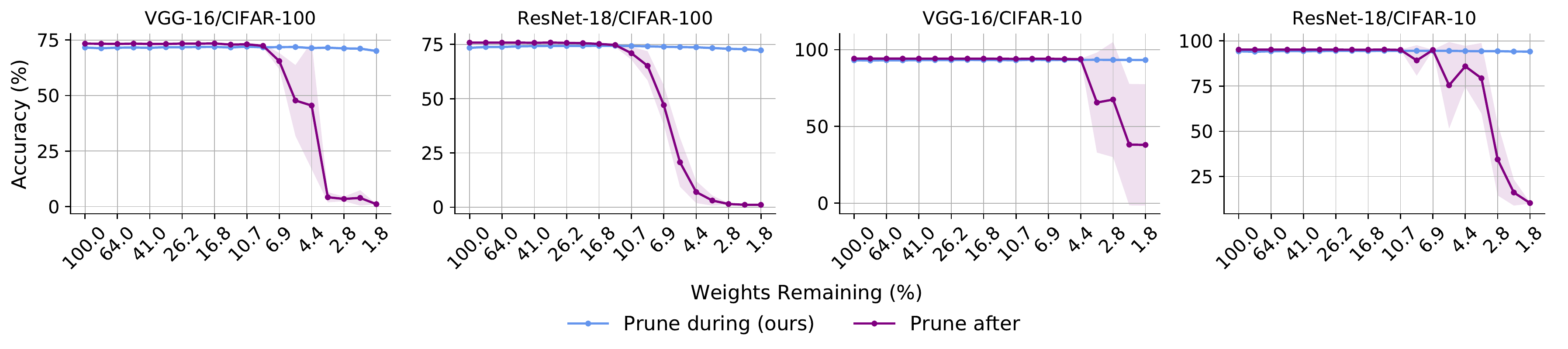} 
\vspace{-1.3em}
\caption{Comparison between prune after interpolating and prune during interpolating.}

\label{fig:prune_time}

\end{figure*}

\begin{figure*}[ht]
\centering
\includegraphics[width=0.99\textwidth]{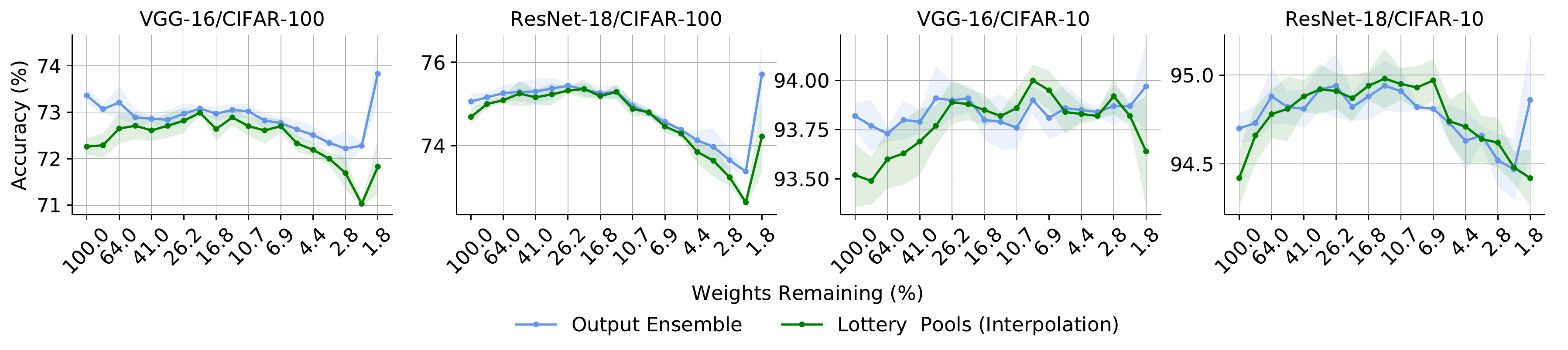} 
\vspace{-0.5em}
\caption{Comparison between Lottery Pools and the output ensemble.}

\label{fig:ensemble}
\end{figure*}

\section{Extensive Analysis}

\textbf{Candidate lottery pools count.} In this section, we study how the number of candidate tickets count, i.e., $\|{S}_t\|_0$ affects the achieved model's performance. In the main experiment section, we take advantage of all the learned networks across different IMP iterations for interpolation. Here, we alter the count of networks in ${S}_t$  as 4, 9, 19, which represents 25\%, 50\%, 100\% total number of  possible networks. All the networks are still sorted in the order of the adjacency to the target lottery tickets in ${S}_t$. The results are shown in Figure~\ref{fig:model_count}. Not surprisingly, more candidate tickets tend to yield better performance in general. Besides, the performance gap between different Candidate Lottery Pools counts in the dense model is more significant than in the sparse situations. 



\textbf{Candidate interpolation coefficient count.} Here, we study how the interpolation coefficient count affects the Lottery Pool's performance. Intentionally, more candidates would be more likely to provide a proper coefficient by searching and thereby gain larger performance improvements. To confirm this hypothesis, we compare Lottery Pools using different candidate coefficient counts, including 1 ($\alpha$ =[0.5]), 3 ($\alpha$=[0.05, 0.5, 0.95]), 7 ($\alpha$=[0.05,0.1,0.3,0.5,0.7,0.9,0.95]). Therefore, when using coefficient count as 1, Lottery Pools scales to Lottery Pools (Average). The results are illustrated in Figure~\ref{fig:model_count}. Lottery Pools (Average) with 1 candidate (yellow lines) achieves the lowest accuracy in general, compared to the settings with more candidates, due to its limited search space.  Lottery Pools with 7 candidates outperforms the one with 3 candidates, but only with marginal gains.    

\textbf{When to Prune.} We study when to prune the interpolated subnetwork to target sparsity. In the default setting of Lottery Pools, we perform the pruning operation every once when we interpolate a subnetwork during the greedy search (line 5, 6 in Algorithm~\ref{alg:Aeraging_recipe}), which we called \textit{prune during}. Another option would be first greedy interpolating all the searched subnetworks and then pruning the achieved subnetwork to the target sparsity, namely \textit{prune after}. We compare these two methods in Figure~\ref{fig:prune_time}. As we can see, the performance of \textit{prune after} drops dramatically at the high sparsity while \textit{prune during} keeps a steady high accuracy across all the sparsities. The reason might be in \textit{prune after}, we greedy search for the best un-pruned interpolated networks regardless of their performance at the target sparsity. Whereas the \textit{prune during} operation in greedy search keeps the interpolated networks always having good performances at desirable sparsity level.

\textbf{Comparison with output ensemble.}
This section compares the performance of Lottery Pools with the output ensemble, i.e. averaging the digit of various subnetworks for inference~\citep{huang2017snapshot,garipov2018loss}. For every learned subnetwork from IMP, we collect the other two identified subnetworks from most adjacent iterations to perform the Lottery Pools and output ensemble. The results are reported in Figure~\ref{fig:ensemble}. As we can see, our Lottery Pools could match the performance of the output ensemble, with no additional computational cost or memory relative to a single subnetwork during inference. 

\section{Conclusion}

In this paper, we explore a new perspective to leverage the existing learned LTs subnetworks by interpolation. We call this approach \textbf{\textit{Lottery Pools}}.   Without increasing training or inference costs, a network can be identified with significant performance improvements for in- and out-distribution scenarios. Impressively, Lottery Pools is capable of creating not only stronger subnetworks that maintain the original LTs sparsity level but also stronger dense networks. Extensive experiments verify the effectiveness of Lottery Pools across various network architectures with  VGG-16 and ResNet-18/34 on CIFAR-10/100 and ImageNet.   

\section{Acknowledgments}
This work used the Dutch national e-infrastructure with the support of the SURF Cooperative using grant no. NWO-2021.060, EINF-2694 and EINF-2943/L1.

\bibliography{aaai23}

\begin{thebibliography}{58}
\providecommand{\natexlab}[1]{#1}

\bibitem[{Breiman(1996)}]{breiman1996bagging}
Breiman, L. 1996.
\newblock Bagging predictors.
\newblock \emph{Machine learning}, 24(2): 123--140.

\bibitem[{Brown et~al.(2020)Brown, Mann, Ryder, Subbiah, Kaplan, Dhariwal,
  Neelakantan, Shyam, Sastry, Askell, Agarwal, Herbert-Voss, Krueger, Henighan,
  Child, Ramesh, Ziegler, Wu, Winter, Hesse, Chen, Sigler, Litwin, Gray, Chess,
  Clark, Berner, McCandlish, Radford, Sutskever, and
  Amodei}]{brown2020language}
Brown, T.; Mann, B.; Ryder, N.; Subbiah, M.; Kaplan, J.~D.; Dhariwal, P.;
  Neelakantan, A.; Shyam, P.; Sastry, G.; Askell, A.; Agarwal, S.;
  Herbert-Voss, A.; Krueger, G.; Henighan, T.; Child, R.; Ramesh, A.; Ziegler,
  D.; Wu, J.; Winter, C.; Hesse, C.; Chen, M.; Sigler, E.; Litwin, M.; Gray,
  S.; Chess, B.; Clark, J.; Berner, C.; McCandlish, S.; Radford, A.; Sutskever,
  I.; and Amodei, D. 2020.
\newblock Language Models are Few-Shot Learners.
\newblock In Larochelle, H.; Ranzato, M.; Hadsell, R.; Balcan, M.~F.; and Lin,
  H., eds., \emph{Advances in Neural Information Processing Systems},
  volume~33, 1877--1901. Curran Associates, Inc.

\bibitem[{Chen et~al.(2022)Chen, Chen, Ma, Wang, and Wang}]{chen2022coarsening}
Chen, T.; Chen, X.; Ma, X.; Wang, Y.; and Wang, Z. 2022.
\newblock Coarsening the granularity: Towards structurally sparse lottery
  tickets.
\newblock \emph{arXiv preprint arXiv:2202.04736}.

\bibitem[{Chen et~al.(2020)Chen, Frankle, Chang, Liu, Zhang, Wang, and
  Carbin}]{chen2020lottery}
Chen, T.; Frankle, J.; Chang, S.; Liu, S.; Zhang, Y.; Wang, Z.; and Carbin, M.
  2020.
\newblock The lottery ticket hypothesis for pre-trained bert networks.
\newblock \emph{Advances in neural information processing systems}, 33:
  15834--15846.

\bibitem[{Chen et~al.(2021)Chen, Zhang, Sui, and Chen}]{chen2021gans}
Chen, X.; Zhang, Z.; Sui, Y.; and Chen, T. 2021.
\newblock Gans can play lottery tickets too.
\newblock \emph{arXiv preprint arXiv:2106.00134}.

\bibitem[{Dietterich(2000)}]{Dietterich2000EnsembleMI}
Dietterich, T.~G. 2000.
\newblock Ensemble Methods in Machine Learning.
\newblock In \emph{Multiple Classifier Systems}.

\bibitem[{Dosovitskiy et~al.(2020)Dosovitskiy, Beyer, Kolesnikov, Weissenborn,
  Zhai, Unterthiner, Dehghani, Minderer, Heigold, Gelly
  et~al.}]{dosovitskiy2020image}
Dosovitskiy, A.; Beyer, L.; Kolesnikov, A.; Weissenborn, D.; Zhai, X.;
  Unterthiner, T.; Dehghani, M.; Minderer, M.; Heigold, G.; Gelly, S.; et~al.
  2020.
\newblock An image is worth 16x16 words: Transformers for image recognition at
  scale.
\newblock \emph{arXiv preprint arXiv:2010.11929}.

\bibitem[{Evci et~al.(2022)Evci, Ioannou, Keskin, and
  Dauphin}]{evci2022gradient}
Evci, U.; Ioannou, Y.; Keskin, C.; and Dauphin, Y. 2022.
\newblock Gradient flow in sparse neural networks and how lottery tickets win.
\newblock In \emph{Proceedings of the AAAI Conference on Artificial
  Intelligence}, volume~36, 6577--6586.

\bibitem[{Fedus, Zoph, and Shazeer(2021)}]{fedus2021switch}
Fedus, W.; Zoph, B.; and Shazeer, N. 2021.
\newblock Switch transformers: Scaling to trillion parameter models with simple
  and efficient sparsity.
\newblock \emph{arXiv preprint arXiv:2101.03961}.

\bibitem[{Fort, Hu, and Lakshminarayanan(2019)}]{fort2019deep}
Fort, S.; Hu, H.; and Lakshminarayanan, B. 2019.
\newblock Deep ensembles: A loss landscape perspective.
\newblock \emph{arXiv preprint arXiv:1912.02757}.

\bibitem[{Frankle and Carbin(2018)}]{frankle2018lottery}
Frankle, J.; and Carbin, M. 2018.
\newblock The lottery ticket hypothesis: Finding sparse, trainable neural
  networks.
\newblock \emph{arXiv preprint arXiv:1803.03635}.

\bibitem[{Frankle et~al.(2020)Frankle, Dziugaite, Roy, and
  Carbin}]{frankle2020linear}
Frankle, J.; Dziugaite, G.~K.; Roy, D.; and Carbin, M. 2020.
\newblock Linear mode connectivity and the lottery ticket hypothesis.
\newblock In \emph{International Conference on Machine Learning}, 3259--3269.
  PMLR.

\bibitem[{Gale, Elsen, and Hooker(2019)}]{gale2019state}
Gale, T.; Elsen, E.; and Hooker, S. 2019.
\newblock The state of sparsity in deep neural networks.
\newblock \emph{arXiv preprint arXiv:1902.09574}.

\bibitem[{Garc{\'\i}a-Mart{\'\i}n et~al.(2019)Garc{\'\i}a-Mart{\'\i}n,
  Rodrigues, Riley, and Grahn}]{garcia2019estimation}
Garc{\'\i}a-Mart{\'\i}n, E.; Rodrigues, C.~F.; Riley, G.; and Grahn, H. 2019.
\newblock Estimation of energy consumption in machine learning.
\newblock \emph{Journal of Parallel and Distributed Computing}, 134: 75--88.

\bibitem[{Garipov et~al.(2018)Garipov, Izmailov, Podoprikhin, Vetrov, and
  Wilson}]{garipov2018loss}
Garipov, T.; Izmailov, P.; Podoprikhin, D.; Vetrov, D.; and Wilson, A.~G. 2018.
\newblock Loss surfaces, mode connectivity, and fast ensembling of dnns.
\newblock In \emph{Proceedings of the 32nd International Conference on Neural
  Information Processing Systems}, 8803--8812.

\bibitem[{Gustafsson, Danelljan, and Schon(2020)}]{gustafsson2020evaluating}
Gustafsson, F.~K.; Danelljan, M.; and Schon, T.~B. 2020.
\newblock Evaluating scalable bayesian deep learning methods for robust
  computer vision.
\newblock In \emph{Proceedings of the IEEE/CVF Conference on Computer Vision
  and Pattern Recognition Workshops}, 318--319.

\bibitem[{Han, Mao, and Dally(2015)}]{han2015deep}
Han, S.; Mao, H.; and Dally, W.~J. 2015.
\newblock Deep compression: Compressing deep neural networks with pruning,
  trained quantization and huffman coding.
\newblock \emph{arXiv preprint arXiv:1510.00149}.

\bibitem[{Hansen and Salamon(1990)}]{hansen1990neural}
Hansen, L.~K.; and Salamon, P. 1990.
\newblock Neural network ensembles.
\newblock \emph{IEEE transactions on pattern analysis and machine
  intelligence}, 12(10): 993--1001.

\bibitem[{He et~al.(2016)He, Zhang, Ren, and Sun}]{resnet}
He, K.; Zhang, X.; Ren, S.; and Sun, J. 2016.
\newblock Deep Residual Learning for Image Recognition.
\newblock In \emph{2016 IEEE Conference on Computer Vision and Pattern
  Recognition (CVPR)}, 770--778.

\bibitem[{Huang et~al.(2017)Huang, Li, Pleiss, Liu, Hopcroft, and
  Weinberger}]{huang2017snapshot}
Huang, G.; Li, Y.; Pleiss, G.; Liu, Z.; Hopcroft, J.~E.; and Weinberger, K.~Q.
  2017.
\newblock Snapshot ensembles: Train 1, get m for free.
\newblock \emph{arXiv preprint arXiv:1704.00109}.

\bibitem[{Ilharco et~al.(2022)Ilharco, Wortsman, Gadre, Song, Hajishirzi,
  Kornblith, Farhadi, and Schmidt}]{patching}
Ilharco, G.; Wortsman, M.; Gadre, S.~Y.; Song, S.; Hajishirzi, H.; Kornblith,
  S.; Farhadi, A.; and Schmidt, L. 2022.
\newblock Patching open-vocabulary models by interpolating weights.

\bibitem[{Izmailov et~al.(2018)Izmailov, Podoprikhin, Garipov, Vetrov, and
  Wilson}]{izmailov2018averaging}
Izmailov, P.; Podoprikhin, D.; Garipov, T.; Vetrov, D.; and Wilson, A.~G. 2018.
\newblock Averaging weights leads to wider optima and better generalization.
\newblock \emph{arXiv preprint arXiv:1803.05407}.

\bibitem[{Jumper et~al.(2021)Jumper, Evans, Pritzel, Green, Figurnov,
  Ronneberger, Tunyasuvunakool, Bates, {\v{Z}}{\'\i}dek, Potapenko
  et~al.}]{jumper2021highly}
Jumper, J.; Evans, R.; Pritzel, A.; Green, T.; Figurnov, M.; Ronneberger, O.;
  Tunyasuvunakool, K.; Bates, R.; {\v{Z}}{\'\i}dek, A.; Potapenko, A.; et~al.
  2021.
\newblock Highly accurate protein structure prediction with AlphaFold.
\newblock \emph{Nature}, 596(7873): 583--589.

\bibitem[{Karras et~al.(2017)Karras, Aila, Laine, and
  Lehtinen}]{karras2017progressive}
Karras, T.; Aila, T.; Laine, S.; and Lehtinen, J. 2017.
\newblock Progressive growing of gans for improved quality, stability, and
  variation.
\newblock \emph{arXiv preprint arXiv:1710.10196}.

\bibitem[{Kingma and Ba(2014)}]{kingma2014adam}
Kingma, D.~P.; and Ba, J. 2014.
\newblock Adam: A method for stochastic optimization.
\newblock \emph{arXiv preprint arXiv:1412.6980}.

\bibitem[{Krizhevsky, Sutskever, and Hinton(2012)}]{Alexnet}
Krizhevsky, A.; Sutskever, I.; and Hinton, G.~E. 2012.
\newblock ImageNet Classification with Deep Convolutional Neural Networks.
\newblock In Pereira, F.; Burges, C. J.~C.; Bottou, L.; and Weinberger, K.~Q.,
  eds., \emph{Advances in Neural Information Processing Systems}, volume~25.
  Curran Associates, Inc.

\bibitem[{Lakshminarayanan, Pritzel, and
  Blundell(2017)}]{lakshminarayanan2017simple}
Lakshminarayanan, B.; Pritzel, A.; and Blundell, C. 2017.
\newblock Simple and scalable predictive uncertainty estimation using deep
  ensembles.
\newblock \emph{Advances in neural information processing systems}, 30.

\bibitem[{LeCun et~al.(1989)LeCun, Boser, Denker, Henderson, Howard, Hubbard,
  and Jackel}]{lecun1989handwritten}
LeCun, Y.; Boser, B.; Denker, J.; Henderson, D.; Howard, R.; Hubbard, W.; and
  Jackel, L. 1989.
\newblock Handwritten digit recognition with a back-propagation network.
\newblock \emph{Advances in neural information processing systems}, 2.

\bibitem[{Levin, Tishby, and Solla(1990)}]{levin1990statistical}
Levin, E.; Tishby, N.; and Solla, S.~A. 1990.
\newblock A statistical approach to learning and generalization in layered
  neural networks.
\newblock \emph{Proceedings of the IEEE}, 78(10): 1568--1574.

\bibitem[{Liu et~al.(2021)Liu, Chen, Atashgahi, Chen, Sokar, Mocanu,
  Pechenizkiy, Wang, and Mocanu}]{liu2021deep}
Liu, S.; Chen, T.; Atashgahi, Z.; Chen, X.; Sokar, G.; Mocanu, E.; Pechenizkiy,
  M.; Wang, Z.; and Mocanu, D.~C. 2021.
\newblock Deep ensembling with no overhead for either training or testing: The
  all-round blessings of dynamic sparsity.
\newblock \emph{arXiv preprint arXiv:2106.14568}.

\bibitem[{Liu et~al.(2022)Liu, Chen, Chen, Shen, Mocanu, Wang, and
  Pechenizkiy}]{liu2022unreasonable}
Liu, S.; Chen, T.; Chen, X.; Shen, L.; Mocanu, D.~C.; Wang, Z.; and
  Pechenizkiy, M. 2022.
\newblock The unreasonable effectiveness of random pruning: Return of the most
  naive baseline for sparse training.
\newblock \emph{arXiv preprint arXiv:2202.02643}.

\bibitem[{Loshchilov and Hutter(2017)}]{loshchilov2017decoupled}
Loshchilov, I.; and Hutter, F. 2017.
\newblock Decoupled weight decay regularization.
\newblock \emph{arXiv preprint arXiv:1711.05101}.

\bibitem[{Molchanov et~al.(2016)Molchanov, Tyree, Karras, Aila, and
  Kautz}]{molchanov2016pruning}
Molchanov, P.; Tyree, S.; Karras, T.; Aila, T.; and Kautz, J. 2016.
\newblock Pruning convolutional neural networks for resource efficient
  inference.
\newblock \emph{arXiv preprint arXiv:1611.06440}.

\bibitem[{Morcos et~al.(2019)Morcos, Yu, Paganini, and Tian}]{morcos2019one}
Morcos, A.; Yu, H.; Paganini, M.; and Tian, Y. 2019.
\newblock One ticket to win them all: generalizing lottery ticket
  initializations across datasets and optimizers.
\newblock \emph{Advances in neural information processing systems}, 32.

\bibitem[{Mozer and Smolensky(1989)}]{mozer1989using}
Mozer, M.~C.; and Smolensky, P. 1989.
\newblock Using relevance to reduce network size automatically.
\newblock \emph{Connection Science}, 1(1): 3--16.

\bibitem[{Nagarajan and Kolter(2019)}]{nagarajan2019uniform}
Nagarajan, V.; and Kolter, J.~Z. 2019.
\newblock Uniform convergence may be unable to explain generalization in deep
  learning.
\newblock \emph{Advances in Neural Information Processing Systems}, 32.

\bibitem[{Neyshabur, Sedghi, and Zhang(2020)}]{neyshabur2020being}
Neyshabur, B.; Sedghi, H.; and Zhang, C. 2020.
\newblock What is being transferred in transfer learning?
\newblock \emph{Advances in neural information processing systems}, 33:
  512--523.

\bibitem[{Ovadia et~al.(2019)Ovadia, Fertig, Ren, Nado, Sculley, Nowozin,
  Dillon, Lakshminarayanan, and Snoek}]{Ovadia2019CanYT}
Ovadia, Y.; Fertig, E.; Ren, J.; Nado, Z.; Sculley, D.; Nowozin, S.; Dillon,
  J.; Lakshminarayanan, B.; and Snoek, J. 2019.
\newblock Can you trust your model\textquotesingle s uncertainty? Evaluating
  predictive uncertainty under dataset shift.
\newblock In Wallach, H.; Larochelle, H.; Beygelzimer, A.; d\textquotesingle
  Alch\'{e}-Buc, F.; Fox, E.; and Garnett, R., eds., \emph{Advances in Neural
  Information Processing Systems}, volume~32. Curran Associates, Inc.

\bibitem[{Patterson et~al.(2021)Patterson, Gonzalez, Le, Liang, Munguia,
  Rothchild, So, Texier, and Dean}]{patterson2021carbon}
Patterson, D.; Gonzalez, J.; Le, Q.; Liang, C.; Munguia, L.-M.; Rothchild, D.;
  So, D.; Texier, M.; and Dean, J. 2021.
\newblock Carbon emissions and large neural network training.
\newblock \emph{arXiv preprint arXiv:2104.10350}.

\bibitem[{Perrone and Cooper(1992)}]{perrone1992networks}
Perrone, M.~P.; and Cooper, L.~N. 1992.
\newblock When networks disagree: Ensemble methods for hybrid neural networks.
\newblock Technical report, Brown Univ Providence Ri Inst for Brain and Neural
  Systems.

\bibitem[{Polyak and Juditsky(1992)}]{polyak1992acceleration}
Polyak, B.~T.; and Juditsky, A.~B. 1992.
\newblock Acceleration of stochastic approximation by averaging.
\newblock \emph{SIAM journal on control and optimization}, 30(4): 838--855.

\bibitem[{Radford et~al.(2021)Radford, Kim, Hallacy, Ramesh, Goh, Agarwal,
  Sastry, Askell, Mishkin, Clark et~al.}]{radford2021learning}
Radford, A.; Kim, J.~W.; Hallacy, C.; Ramesh, A.; Goh, G.; Agarwal, S.; Sastry,
  G.; Askell, A.; Mishkin, P.; Clark, J.; et~al. 2021.
\newblock Learning transferable visual models from natural language
  supervision.
\newblock \emph{arXiv preprint arXiv:2103.00020}.

\bibitem[{Ramanujan et~al.(2020)Ramanujan, Wortsman, Kembhavi, Farhadi, and
  Rastegari}]{ramanujan2020s}
Ramanujan, V.; Wortsman, M.; Kembhavi, A.; Farhadi, A.; and Rastegari, M. 2020.
\newblock What's hidden in a randomly weighted neural network?
\newblock In \emph{Proceedings of the IEEE/CVF Conference on Computer Vision
  and Pattern Recognition}, 11893--11902.

\bibitem[{Rame et~al.(2022)Rame, Kirchmeyer, Rahier, Rakotomamonjy, Gallinari,
  and Cord}]{rame2022diverse}
Rame, A.; Kirchmeyer, M.; Rahier, T.; Rakotomamonjy, A.; Gallinari, P.; and
  Cord, M. 2022.
\newblock Diverse Weight Averaging for Out-of-Distribution Generalization.
\newblock \emph{arXiv preprint arXiv:2205.09739}.

\bibitem[{Renda, Frankle, and Carbin(2020)}]{renda2020comparing}
Renda, A.; Frankle, J.; and Carbin, M. 2020.
\newblock Comparing rewinding and fine-tuning in neural network pruning.
\newblock \emph{arXiv preprint arXiv:2003.02389}.

\bibitem[{Ruppert(1988)}]{ruppert1988efficient}
Ruppert, D. 1988.
\newblock Efficient estimations from a slowly convergent Robbins-Monro process.
\newblock Technical report, Cornell University Operations Research and
  Industrial Engineering.

\bibitem[{Schwartz et~al.(2020)Schwartz, Dodge, Smith, and
  Etzioni}]{schwartz2020green}
Schwartz, R.; Dodge, J.; Smith, N.~A.; and Etzioni, O. 2020.
\newblock Green ai.
\newblock \emph{Communications of the ACM}, 63(12): 54--63.

\bibitem[{Silver et~al.(2016)Silver, Huang, Maddison, Guez, Sifre, Van
  Den~Driessche, Schrittwieser, Antonoglou, Panneershelvam, Lanctot
  et~al.}]{silver2016mastering}
Silver, D.; Huang, A.; Maddison, C.~J.; Guez, A.; Sifre, L.; Van Den~Driessche,
  G.; Schrittwieser, J.; Antonoglou, I.; Panneershelvam, V.; Lanctot, M.;
  et~al. 2016.
\newblock Mastering the game of Go with deep neural networks and tree search.
\newblock \emph{nature}, 529(7587): 484--489.

\bibitem[{Simonyan and Zisserman(2014)}]{simonyan2014very}
Simonyan, K.; and Zisserman, A. 2014.
\newblock Very deep convolutional networks for large-scale image recognition.
\newblock \emph{arXiv preprint arXiv:1409.1556}.

\bibitem[{Sun et~al.(2021)Sun, Qin, Zhang, Ma, Li, Luo, Zhao, Chen, and
  Xie}]{sun2021load}
Sun, F.; Qin, M.; Zhang, T.; Ma, X.; Li, H.; Luo, J.; Zhao, Z.; Chen, Y.-K.;
  and Xie, Y. 2021.
\newblock Load-balanced Gather-scatter Patterns for Sparse Deep Neural
  Networks.
\newblock \emph{arXiv preprint arXiv:2112.10898}.

\bibitem[{Wen, Tran, and Ba(2020)}]{wen2020batchensemble}
Wen, Y.; Tran, D.; and Ba, J. 2020.
\newblock BatchEnsemble: an Alternative Approach to Efficient Ensemble and
  Lifelong Learning.
\newblock In \emph{International Conference on Learning Representations}.

\bibitem[{Wortsman et~al.(2022)Wortsman, Ilharco, Gadre, Roelofs,
  Gontijo-Lopes, Morcos, Namkoong, Farhadi, Carmon, Kornblith
  et~al.}]{wortsman2022model}
Wortsman, M.; Ilharco, G.; Gadre, S.~Y.; Roelofs, R.; Gontijo-Lopes, R.;
  Morcos, A.~S.; Namkoong, H.; Farhadi, A.; Carmon, Y.; Kornblith, S.; et~al.
  2022.
\newblock Model soups: averaging weights of multiple fine-tuned models improves
  accuracy without increasing inference time.
\newblock In \emph{International Conference on Machine Learning}, 23965--23998.
  PMLR.

\bibitem[{Yin et~al.(2022)Yin, Menkovski, Fang, Huang, Pei, Pechenizkiy,
  Mocanu, and Liu}]{yin2022superposing}
Yin, L.; Menkovski, V.; Fang, M.; Huang, T.; Pei, Y.; Pechenizkiy, M.; Mocanu,
  D.~C.; and Liu, S. 2022.
\newblock Superposing Many Tickets into One: A Performance Booster for Sparse
  Neural Network Training.
\newblock \emph{arXiv preprint arXiv:2205.15322}.

\bibitem[{Yu et~al.(2019)Yu, Edunov, Tian, and Morcos}]{yu2019playing}
Yu, H.; Edunov, S.; Tian, Y.; and Morcos, A.~S. 2019.
\newblock Playing the lottery with rewards and multiple languages: lottery
  tickets in rl and nlp.
\newblock \emph{arXiv preprint arXiv:1906.02768}.

\bibitem[{Yuan et~al.(2021)Yuan, Ma, Niu, Li, Kong, Liu, Gong, Zhan, He, Jin
  et~al.}]{yuan2021mest}
Yuan, G.; Ma, X.; Niu, W.; Li, Z.; Kong, Z.; Liu, N.; Gong, Y.; Zhan, Z.; He,
  C.; Jin, Q.; et~al. 2021.
\newblock Mest: Accurate and fast memory-economic sparse training framework on
  the edge.
\newblock \emph{Advances in Neural Information Processing Systems}, 34:
  20838--20850.

\bibitem[{Zhang et~al.(2019)Zhang, Lucas, Ba, and Hinton}]{zhang2019lookahead}
Zhang, M.; Lucas, J.; Ba, J.; and Hinton, G.~E. 2019.
\newblock Lookahead optimizer: k steps forward, 1 step back.
\newblock \emph{Advances in neural information processing systems}, 32.

\bibitem[{Zhang et~al.(2022)Zhang, Chen, Xu, Cao, Chen, Cohn, and
  Fang}]{zhang2022survey}
Zhang, Q.; Chen, S.; Xu, D.; Cao, Q.; Chen, X.; Cohn, T.; and Fang, M. 2022.
\newblock A Survey for Efficient Open Domain Question Answering.
\newblock \emph{arXiv preprint arXiv:2211.07886}.

\bibitem[{Zhou et~al.(2019)Zhou, Lan, Liu, and
  Yosinski}]{zhou2019deconstructing}
Zhou, H.; Lan, J.; Liu, R.; and Yosinski, J. 2019.
\newblock Deconstructing lottery tickets: Zeros, signs, and the supermask.
\newblock \emph{Advances in neural information processing systems}, 32.

\end{thebibliography}

\appendix

\renewcommand\thefigure{\Alph{section}\arabic{figure}}  
\onecolumn


\section{Results of ImageNet}
\begin{figure}[htbp]
\centering
{
\includegraphics[width=0.7
\textwidth]{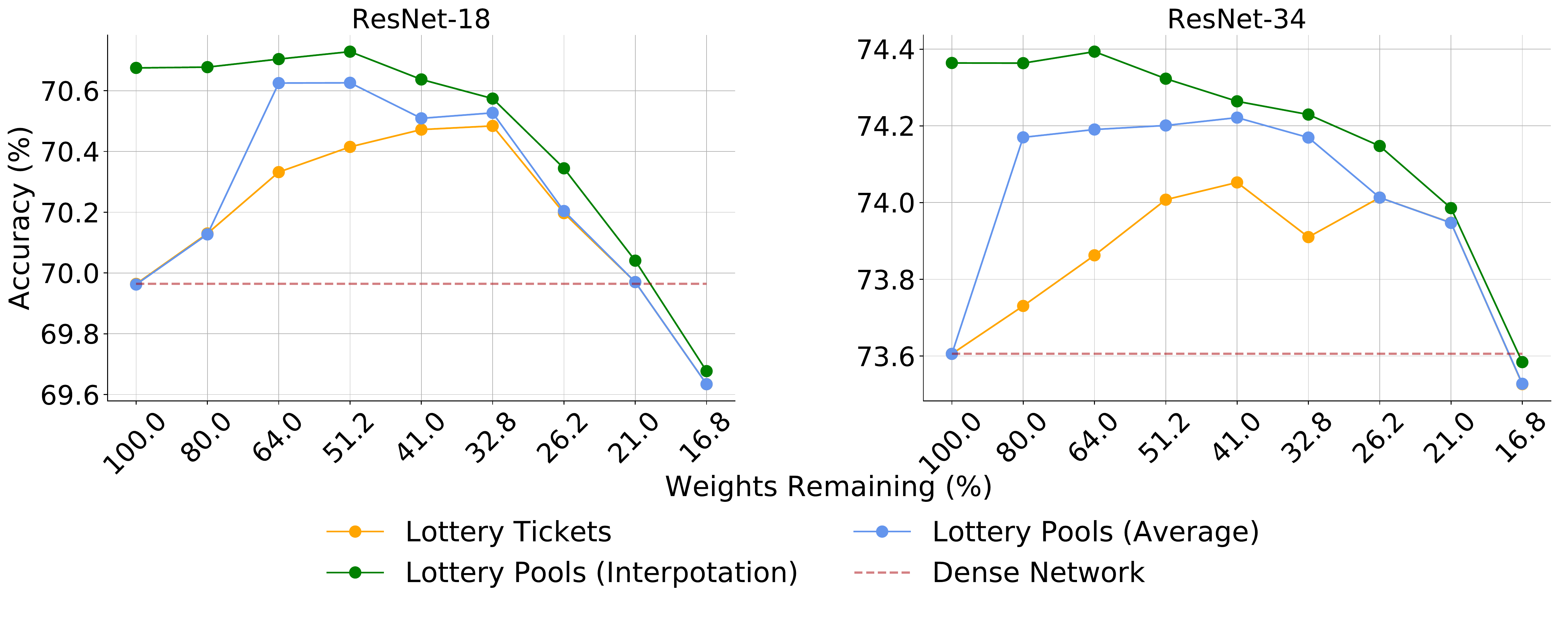} 

\caption{Test accuracy {\%} of original Lottery Tickets and Lottery Pools on ImageNet.}.
\label{fig:Imagenet_acc}}
\end{figure}

In this appendix, we compare the performance of Lottery Pools against the original Lottery Tickets on ImageNet and report the results in Figure\ref{fig:Imagenet_acc}. Overall, we notice a clear performance gain from Lottery Pools over the original LTs at a set of sparsities.

To be specific, by adopting Lottery Pools (Interpolation),  we observe up to 0.55\% and  0.63\% improvements to the original sparse LTs on ResNet-18 and ResNet-34, respectively.  Besides, the constructed stronger dense networks outperform the original dense  ResNet-18 and ResNet-34 by 0.71\%  and 0.76\%, respectively.  All these results demonstrated  Lottery Pools' effectiveness on the large-scale dataset.

\section{Accuracy Heat Map of VGG-16 }
\begin{figure}[ht]
\centering
    \subfigure[VGG-16 w/ Rewinding]{
        \includegraphics[width=0.3\textwidth]{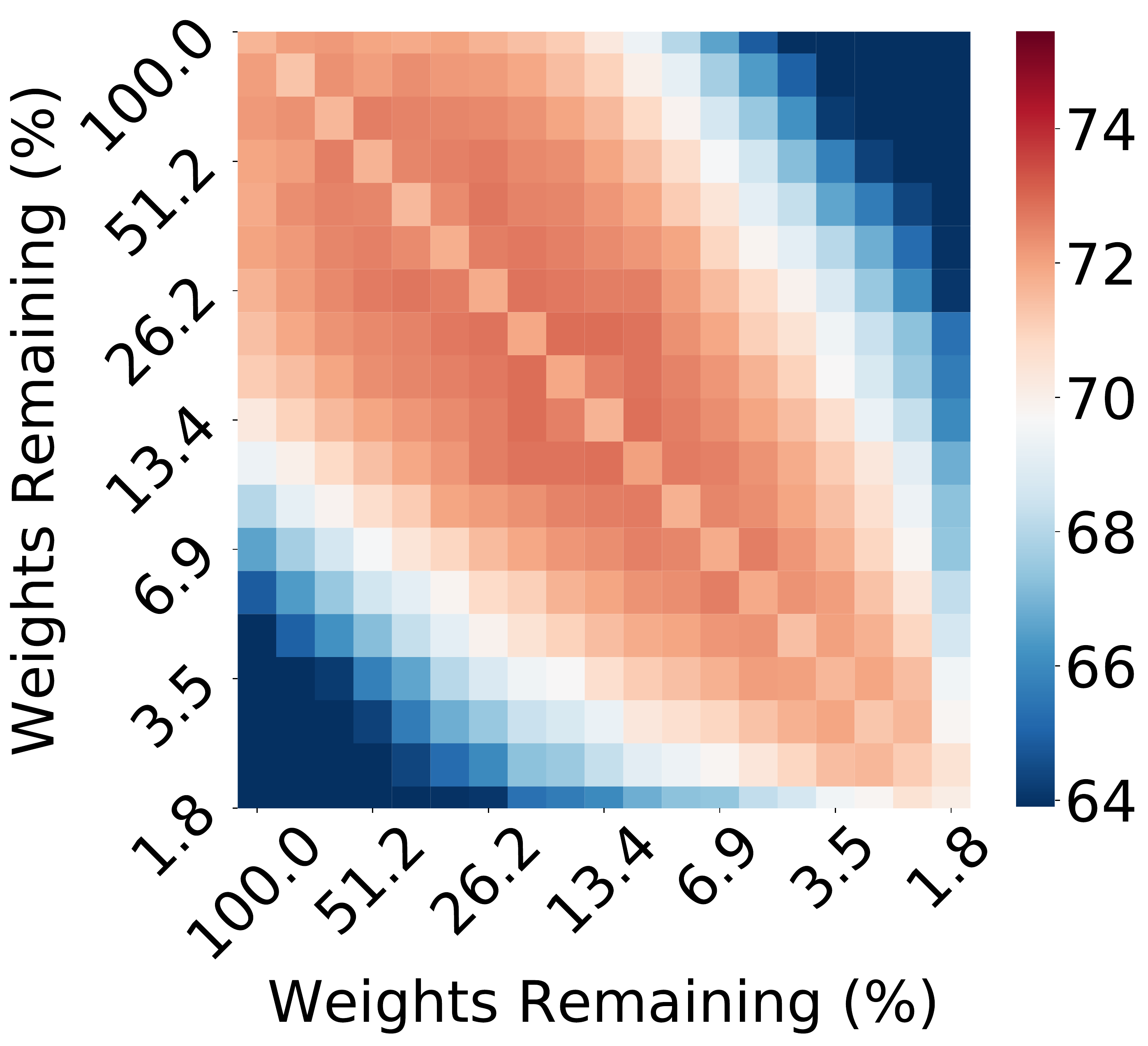}
    }
     \subfigure[VGG-16 w/o Rewinding]{
        \includegraphics[width=0.3\textwidth]{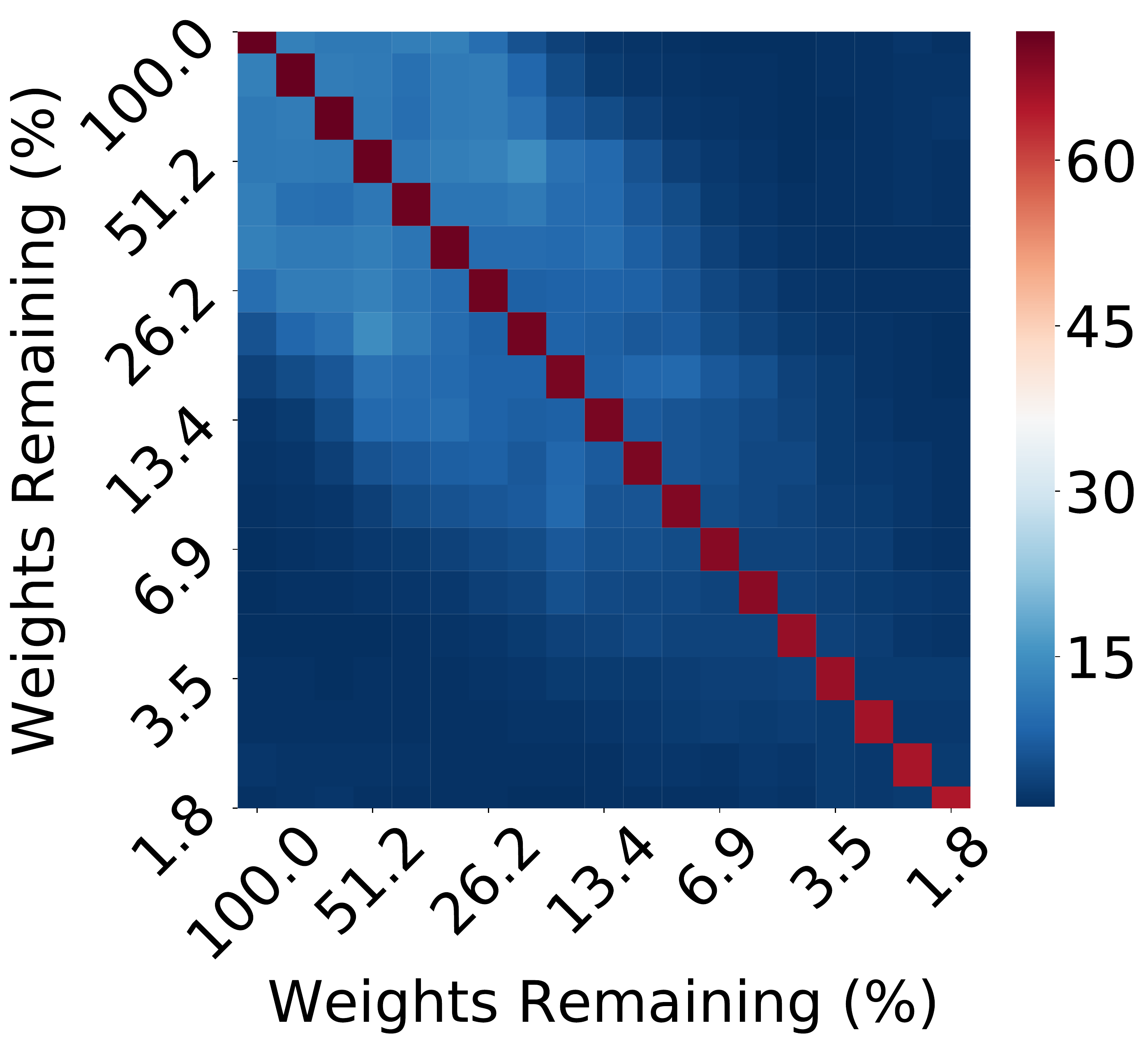}
    }

\caption{Accuracy heat map of the averaged LTs on CIFAR-100. Each cell refers to the test accuracy of the averaged subnetworks using the LTs under the sparsity of X-axis and Y-axis. If the averaged subnetwork decreases in sparsity, we prune it to the higher sparsity of its parents.}
    \label{fig:VGG_ACC_heatmap}

\end{figure}

The accuracy heat map of VGG-16 is reported in Figure~\ref{fig:VGG_ACC_heatmap}. Here, we observe a similar pattern to the results of Resnet-18 (Figure~\ref{fig:ACC_heatmap} in the main paper). First, we notice that rewinding is necessary for performance gain using weight averaging. Secondly, the averaged subnetwork's performance highly depends on its parents' IMP iteration adjacency. The closer IMP iterations two LTs are from, the better their averaged subnetwork performs. Under rewinding,  weight averaging could achieve higher accuracy than the original  LTs subnetwork at the same sparsity level if the parent subnetworks are close enough in IMP iteration.

\clearpage
\newpage

\section{Candidate Interpolation Coefficients}

 In Table~\ref{tab:Coefficient}, we report the Candidate Interpolation Coefficients used in the main experiments of this work. We apply 12 candidate coefficient values ranging from 0.05 to 0.95 to enlarge the optimal value searching space. In Table~\ref{tab:Coefficient_extensive}, we show the values adopted in the chapter ``Extensive Analysis'' of the main paper, where the effect of interpolation coefficient count on Lottery Pools' performance is studied.

\begin{table}[htbp]
    \centering

    \caption{Candidate interpolation coefficient in main experiments.}
    \vspace{-1em}
    \label{tab:Coefficient}
    \resizebox{0.6\textwidth}{!}{
    \begin{tabular}{l|l|c|c}
        \toprule
         Network  & Dataset & Coefficient Count & Coefficient Values    \\ 
         \cmidrule(lr){1-2}
         \cmidrule(lr){3-3}
         \cmidrule(lr){4-4}
         ResNet-18/VGG-16 & CIFAR-10/100  & \multirow{2}{*}{11}  & {0.05, 0.1, 0.2, 0.3, 0.4, 0.5}\\
         ResNet-18/34& ImageNet  &    & {0.6, 0.7, 0.8, 0.9, 0.95} \\

         \bottomrule
    \end{tabular}}
\end{table}

\begin{table}[htbp]
    \centering

    \caption{Candidate interpolation coefficient in  extensive analysis.}
    \vspace{-1em}
    \label{tab:Coefficient_extensive}
    \resizebox{0.6\textwidth}{!}{
    \begin{tabular}{l|l|c|c}
        \toprule
         Network  & Dataset & Coefficient Count & Coefficient Values    \\ 
         \cmidrule(lr){1-2}
         \cmidrule(lr){3-3}
         \cmidrule(lr){4-4}
         ResNet-18/VGG-16 & CIFAR-10/100  & {1}  & 
         {0.5}   \\
         \cmidrule(lr){1-2}
         \cmidrule(lr){3-3}
         \cmidrule(lr){4-4}
         
         ResNet-18/VGG-16 & CIFAR-10/100  & {3}  & 
         {0.05, 0.5, 0.95}   \\

         \cmidrule(lr){1-2}
         \cmidrule(lr){3-3}
         \cmidrule(lr){4-4}
         
         ResNet-18/VGG-16 & CIFAR-10/100  & {7}  & 
         {0.05, 0.1, 0.3, 0.5, 0.7, 0.9, 0.95}   \\
         
         
         
         \bottomrule
    \end{tabular}}
\end{table}

\section{Dataset Details}

We reported the datasets details, including the number of classes, the size of training, validation and testing sets on CIFAR-10, CIFAR-100, and ImageNet in Table~\ref{tab:data}. As there are no publicly available labeled sets for testing in these datasets, we use original validation sets for testing, split  10\% of the training sets as hold-out validation sets, and use the rest of the training sets for training.

\begin{table*}[htbp]
    \centering
    \caption{Datasets Details.}
    \vspace{-1em}
    \label{tab:data}
    \resizebox{0.5\textwidth}{!}{
    \begin{tabular}{lrrrr}
    \toprule
    & \multicolumn{3}{c}{Size of the set used for} & \\
    Dataset & Training & Validation & Testing & Number of classes\\
    \midrule
    CIFAR-10 & 45,000 & 5,000 & 10,000 & 10\\
    CIFAR-100  & 45,000 & 5,000 & 10,000 & 100\\
    ImageNet  & 1,255,167 & 26,000 & 50,000 & 1,000\\
    \bottomrule
    \end{tabular}}

\end{table*}

\section{Lottery Tickets Implementation Details}

 In Table~\ref{tab:hypo_hyper_cifar} and Table~\ref{tab:hypo_hyper_imgnet}, we report the implementation details of creating Lottery Tickets subnetworks that are used for interpolating in Lottery Pools. The reported hyperparameters include total training epochs (Epoch), learning rate (LR), batch size (BS),  learning rate drop (LR Drop), weight decay (WD), SGD momentum (Momentum), IMP iteration count, IMP weight pruning fraction and the rewinding epochs, etc.

\vspace{1em}
\textbf{Computation resources.} The experiments on ImageNet were performed with 4 NVIDIA Tesla A100 GPUs, and the experiments on CIFAR-10/100  were run on a single A100 GPU.

\vspace{1em}
\textbf{Implementation details on CIFAR-10/100.}
.

\begin{table*}[!ht]
\centering
\caption{Implementation hyperparameters of  Lottery Tickets on CIFAR-10/100.}
\label{tab:hypo_hyper_cifar}
\resizebox{1.0\textwidth}{!}{
\begin{tabular}{cccccccccccccccc}
\toprule
Model  & Epoch & BS & LR & LR Drop, Epochs  & Optimizer & WD & Momentum & Warmup (epochs) &  Rewinding (epochs)  &  IMP iterations (epochs) &  Pruning Fraction (\%) \\ 
\toprule
VGG-16 &182  & 128 & 0.1 & 10x, [91, 136]  & SGD &  0.9 & 1e-4 & - & 9   &  19   & 20  \\
ResNet-18  &182  & 128 & 0.1 & 10x, [91, 136]  & SGD &  0.9 & 1e-4 & - & 9   &  19   & 20  \\
\bottomrule
\end{tabular}}
\end{table*}

\textbf{Implementation details  on ImageNet.}

\begin{table*}[!ht]
\centering
\caption{Implementation hyperparameters of Lottery Tickets on ImageNet.}
\label{tab:hypo_hyper_imgnet}
\resizebox{1.0\textwidth}{!}{
\begin{tabular}{cccccccccccccccc}
\toprule
Model  & Epoch & BS & LR & LR Drop, Epochs  & Optimizer & WD & Momentum & Warmup (epochs) &  Rewinding (epochs)  &  IMP iterations (epochs) &  Pruning Fraction (\%) \\ 
\toprule
ResNet-18 & 90  & 1024 & 0.4 & 10x, [30, 60, 80]  & SGD &  0.9 & 1e-4 & 5 & 5   &  9   & 20  \\
ResNet-34 & 90  & 1024 & 0.4 & 10x, [30, 60, 80]  & SGD &  0.9 & 1e-4 & 5 & 5   &  9   & 20  \\
\bottomrule
\end{tabular}}
\end{table*}

\section{Algorithm of Lottery Pools (Average)}

When applying a fixed interpolation value of 0.5 instead of the searched optimal one within the Candidate Coefficient Pools,  the Lottery Pools (Interpolation) turns into its variant, the Lottery Pools (Average) that is described in Algorithm~\ref{alg:Aeraging_recipe_average}.

Lottery Pools (Interpolation) tends to achieve better performance than Lottery Pools (Average) by using the searched optimal value for interpolating, while Lottery Pools (Average)  is benefited from efficiency due to simple averaging.

\begin{algorithm*}[htbp]
  \caption{Lottery Pools Average Recipe}
  \label{alg:Aeraging_recipe_average}
   
\textbf{Input}: The original learned sparse subnetwork ${\bm{\mathrm{\tilde\theta}}}_{t}$ from LTs,  Candidate Lottery Pools $ \mathcal{S}_t = \{  {\bm{\mathrm{\tilde\theta}}_{t\text{-}1}},{\bm{\mathrm{\tilde\theta}}_{t\text{+}1}},{\bm{\mathrm{\tilde\theta}}_{t\text{+}2}},\cdots \}$, The averaged subnetwork ${\bm{\mathrm{\tilde\theta}}_{Inter}}$ during greedy search. 

\textbf{Output}: Final averaged subnetwork ${\bm{\mathrm{\tilde\theta}}_{best}}$ that has the same sparsity with ${\bm{\mathrm{\tilde\theta}}_t}$.
\begin{algorithmic}[1]

    \STATE $ \mathcal{S}_t  \gets \{  {\bm{\mathrm{\tilde\theta}}_{t\text{-}1}},{\bm{\mathrm{\tilde\theta}}_{t\text{+}1}},{\bm{\mathrm{\tilde\theta}}_{t\text{+}2}},\cdots \}$  \COMMENT{ Create Candidate Lottery Pools, and sort all candidates by their adjacence to  ${\bm{\mathrm{\tilde\theta}}_t}$ }

    \STATE ${\bm{\mathrm{\tilde\theta }}_{best}}   \gets  {\bm{\mathrm{\tilde\theta}}}_t$

    \STATE  {\bfseries for} ${\bm{\mathrm{\tilde\theta}}_i}  \in \mathcal{S}_t  $  {\bfseries do}   \COMMENT{ Greedily search the candidate LTs subnetworks for averaging  }

    \STATE \ \ \  ${\bm{\mathrm{\tilde\theta}}_{Inter}} \gets \mathsf{MagnitudePruning}\mleft( \frac{ \bm{\mathrm{\tilde\theta}}_{best} + \bm{\mathrm{\tilde\theta}}_{i} }{2}\mright) $  \COMMENT{Average }

    \STATE  \ \ \ {\bfseries if} $ \mathsf{ValAcc}\mleft({\bm{\mathrm{\tilde\theta}}_{Inter}} \mright)\geq \mathsf{ValAcc} \mleft( {\bm{\mathrm{\tilde\theta}}_{best}}  \mright) $ 
    
    \STATE \ \ \ \ \ \  \ \ \   {\bfseries then}     ${\bm{\mathrm{\tilde\theta}}_{best}} \gets    {\bm{\mathrm{\tilde\theta}}_{Inter}}$ 
    \COMMENT{Update the best averaged subnetwork}
    

    \STATE  {\bfseries end for}

\end{algorithmic}
\end{algorithm*}

\section{Diversity analysis.} 
\begin{figure}[ht]
\centering
    \subfigure[\small{VGG-16/CIFAR-100}]{
        \includegraphics[width=0.3\textwidth]{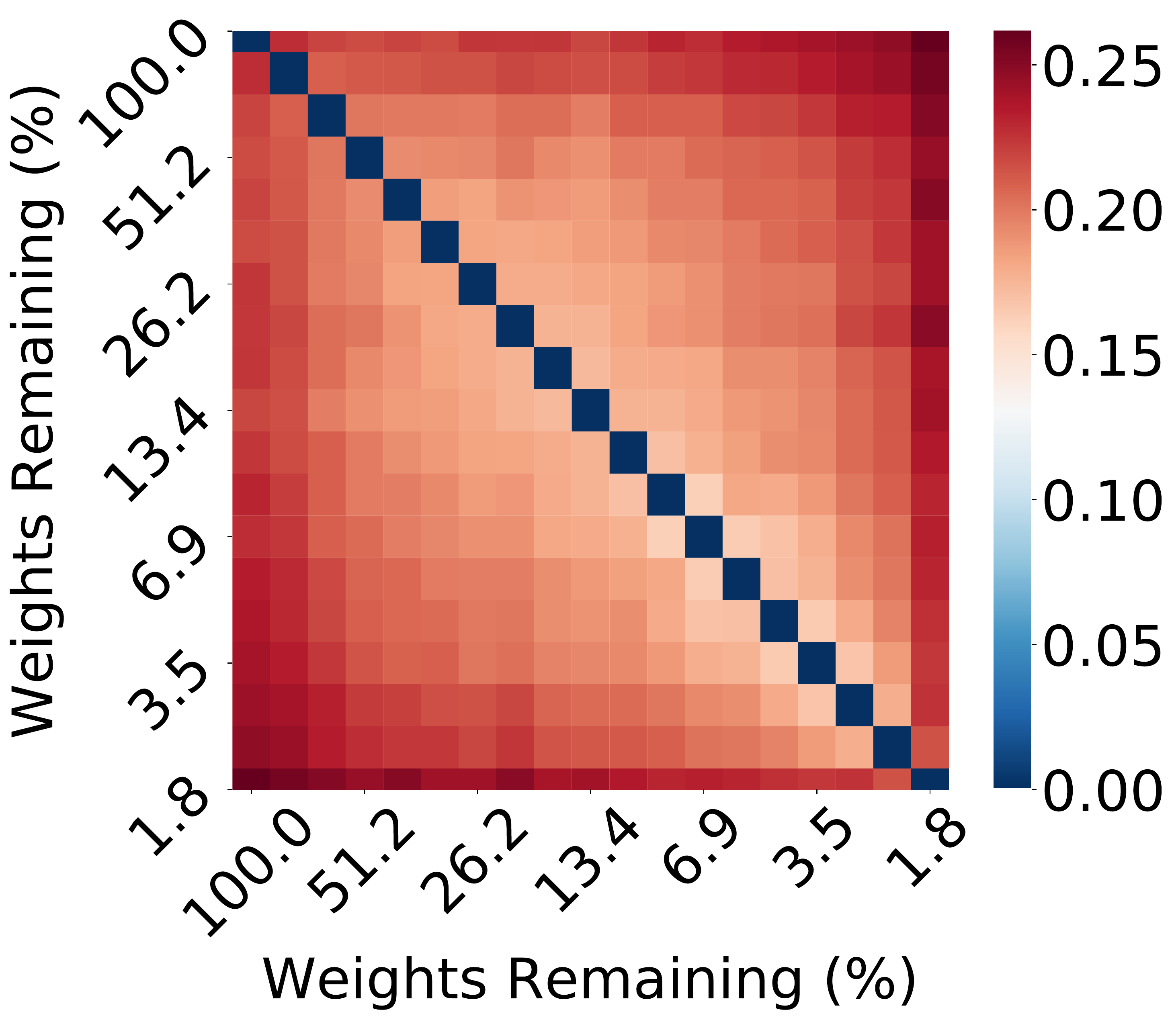}
    }
    \subfigure[ResNet-18/CIFAR-100]{
        \includegraphics[width=0.3\textwidth]{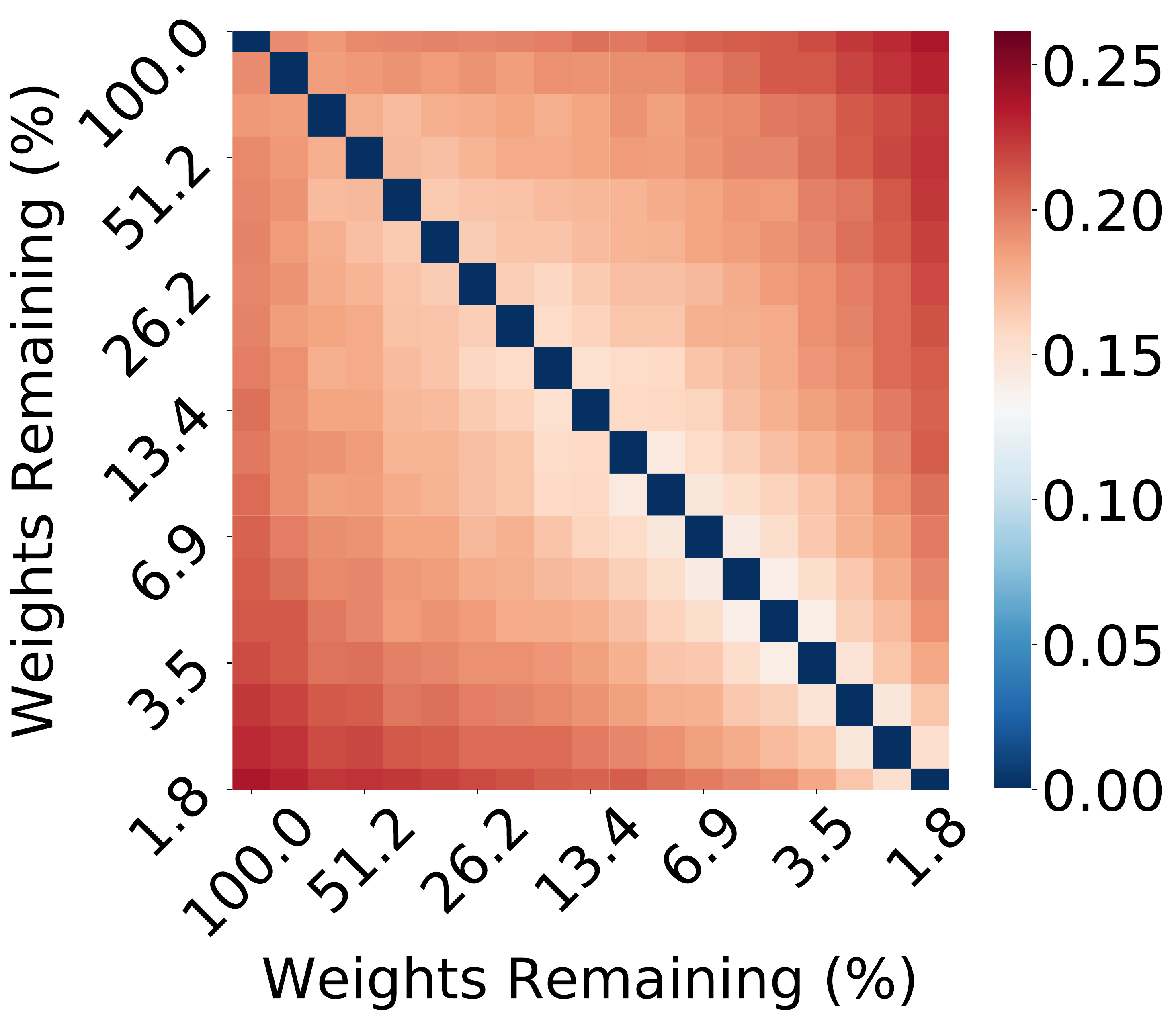}
    }
    
\caption{Prediction disagreement between lottery tickets at the different sparsity levels. Each block in the heatmap shows the fraction of labels on which the predictions from different lottery tickets disagree.}
    \label{fig:disagree_heatmap}

\end{figure}

We plot the prediction disagreement matrix across the learned subnetwork from IMP across different iterations in  Figure~\ref{fig:disagree_heatmap}. We can see that the LTs subnetworks behave similarly to their neighbor subnetworks with small prediction disagreement, and such disagreement gradually increases as their distance becomes larger and larger in the context of IMP. The results could explain the phenomena of Figure~\ref{fig:ACC_heatmap} as a larger diversity indicates it is more likely that the subnetworks are located in the different basins, where weight interpolation does not provide satisfactory performance~\citep{neyshabur2020being,yin2022superposing}.

\clearpage
\newpage

\section{Results Tables}

Here we report the performance details of Lottery Pools, EMA, SWA and original Lottery tickets on CIFAR-10/100 in Table~\ref{tab:cifar_results_1} and Table~\ref{tab:cifar_results_2}. The performance details on ImageNet are reported  in~Table~\ref{tab:imgnet_result}. We apply the decay factor of 0.95 in EMA. For a fair comparison, EMA and SWA are modified by pruning the interpolated models to the sparsity of the original LHs during interpolating. The reported accuracy on CIFAR-10/100 is averaged over 3 independent runs, while we only run the experiments on ImageNet once due to the limited resources.

Compared with the original LTs, Lottery Pools achieves universal performance improvements at all sparsity levels on both ImageNet and CIFAR-10/100 datasets. Compared with the other two baselines (SWA and EMA), our Lottery Pools outperforms these baselines in all cases on ImageNet, and in most cases (60/76) on CIFAR-10/100.

\begin{table*}[htbp]
    \centering
    \caption{Accuracy (\%)  of Lottery Pools,  the original Lottery Tickets, EMA and SWA on CIFAR-10/100  (1).}

    \label{tab:cifar_results_1}
    \resizebox{.8\textwidth}{!}{
    \begin{tabular}{l|c|c| ccccc cccc}
        \hline
        \multirow{2}{*}{Dataset}  &  \multirow{2}{*}{Network} & \multirow{2}{*}{Method}  & \multicolumn{9}{c}{Weights Remaining\%} \\
         \cmidrule(lr){4-12}
        & &   &   100& 80& 64& 51.20& 40.96& 32.77& 26.21& 20.97& 16.78   \\
        \cmidrule(lr){1-1}
        \cmidrule(lr){2-2}
        \cmidrule(lr){3-3}
        \cmidrule(lr){4-12}
        
          \multirow{5}{*}{CIFAR-100}  & \multirow{5}{*}{VGG-16} & Lottery Pools (interpolation)  & \textbf{73.32$\pm$0.22} & \textbf{73.18$\pm$0.38} & {72.82$\pm$0.17} & \textbf{73.27$\pm$0.26} & {73.13$\pm$0.38} & {73.04$\pm$0.20} & \textbf{73.32$\pm$0.24} & {73.35$\pm$0.16} & {73.23$\pm$0.13} \\ 
         
          & & Lottery Pools (Average)   & 73.14$\pm$0.30& 73.11$\pm$0.33 & 72.68$\pm$0.34 & 72.80$\pm$0.30& 72.93$\pm$0.29 & 72.96$\pm$0.35 & 72.97$\pm$0.45 & 72.98$\pm$0.18 & 73.16$\pm$0.04  \\ 
          
          & & SWA   &  73.14$\pm$0.34 & 73.15$\pm$0.33 & \textbf{73.13$\pm$0.33} & 73.15$\pm$0.32 & \textbf{73.23$\pm$0.24} & \textbf{73.26$\pm$0.25} & 73.32$\pm$0.27 & \textbf{73.44$\pm$0.30} & \textbf{73.48$\pm$0.13}  \\ 
          
          & & EMA (0.95)  & 72.23$\pm$0.29 & 72.40$\pm$0.21 & 72.70$\pm$0.13 & 73.05$\pm$0.17 & 72.93$\pm$0.22 & 73.20$\pm$0.38 & 73.23$\pm$0.18 & 73.37$\pm$0.23 & 73.34$\pm$0.05  \\ 
          
        & & Original LTs   & 71.63$\pm$0.19 & 71.30$\pm$0.27 & 71.58$\pm$0.17 & 71.65$\pm$0.23 & 71.53$\pm$0.37 & 71.77$\pm$0.19 & 71.79$\pm$0.25 & 71.88$\pm$0.30& 71.91$\pm$0.18 \\ 

        \cmidrule(lr){1-1}
        \cmidrule(lr){2-2}
        \cmidrule(lr){3-3}
      \cmidrule(lr){4-12}

            \multirow{5}{*}{CIFAR-100}  & \multirow{5}{*}{ResNet-18} & Lottery Pools (interpolation) & \textbf{75.67$\pm$0.30} & {75.54$\pm$0.20} & {75.53$\pm$0.25} & \textbf{75.74$\pm$0.23} & \textbf{75.77$\pm$0.12} & \textbf{75.82$\pm$0.23} & \textbf{75.97$\pm$0.21} & \textbf{75.69$\pm$0.18} & {75.42$\pm$0.07} \\ 
         
          & & Lottery Pools (Average)   &  75.65$\pm$0.13 & {75.58$\pm$0.11} & 75.44$\pm$0.33 & 75.68$\pm$0.18 & 75.65$\pm$0.28 & 75.34$\pm$0.29 & 75.32$\pm$0.18 & 75.58$\pm$0.07 & 75.31$\pm$0.12 \\ 
          
          & & SWA   &   75.60$\pm$0.06 & \textbf{75.62$\pm$0.05} & \textbf{75.62$\pm$0.06} & 75.60$\pm$0.08 & 75.71$\pm$0.06 & 75.72$\pm$0.09 & 75.71$\pm$0.15 & 75.68$\pm$0.12 & {75.58$\pm$0.24} \\ 
          
          & & EMA (0.95)  & 74.49$\pm$0.19 & 74.98$\pm$0.05 & 75.14$\pm$0.13 & 75.28$\pm$0.05 & 75.50$\pm$0.28 & 75.54$\pm$0.10& 75.61$\pm$0.17 & 75.66$\pm$0.10& \textbf{75.59$\pm$0.23}  \\ 
          
        & & Original LTs   & 73.45$\pm$0.27 & 73.82$\pm$0.06 & 73.84$\pm$0.07 & 74.12$\pm$0.24 & 74.24$\pm$0.23 & 74.32$\pm$0.27 & 74.31$\pm$0.20& 74.34$\pm$0.15 & 74.35$\pm$0.22 \\

        \cmidrule(lr){1-1}
        \cmidrule(lr){2-2}
        \cmidrule(lr){3-3}
      \cmidrule(lr){4-12}

           \multirow{5}{*}{CIFAR-10}  & \multirow{5}{*}{VGG-16} & Lottery Pools (interpolation)  & {93.76$\pm$0.18} & {93.90$\pm$0.14} & {93.97$\pm$0.04} & {93.88$\pm$0.03} & {93.83$\pm$0.04 }& \textbf{94.01$\pm$0.15} & \textbf{94.03$\pm$0.14} & \textbf{94.11$\pm$0.03} & {93.98$\pm$0.11} \\ 
         
          & & Lottery Pools (Average)   &  93.74$\pm$0.19 & 93.95$\pm$0.13 & \textbf{94.09$\pm$0.15} & \textbf{94.01$\pm$0.14} & 93.89$\pm$0.16 & 93.92$\pm$0.05 & 93.79$\pm$0.18 & 94.04$\pm$0.08 & 94.04$\pm$0.09 \\ 
          
          & & SWA   &  \textbf{94.01$\pm$0.06} & \textbf{94.01$\pm$0.07} & 94.01$\pm$0.06 & 94.00$\pm$0.06 & \textbf{94.02$\pm$0.08 }& {94.00$\pm$0.02} & 94.01$\pm$0.05 & 94.07$\pm$0.07 & \textbf{94.08$\pm$0.12}  \\ 
          
          & & EMA (0.95)  & 93.65$\pm$0.10& 93.76$\pm$0.14 & 93.76$\pm$0.16 & 93.66$\pm$0.07 & 93.98$\pm$0.09 & 93.88$\pm$0.13 & 94.00$\pm$0.04 & 93.98$\pm$0.07 & 93.97$\pm$0.06  \\ 
          
        & & Original LTs   & 93.14$\pm$0.11 & 92.98$\pm$0.25 & 93.14$\pm$0.15 & 93.10$\pm$0.14 & 93.21$\pm$0.10& 93.25$\pm$0.23 & 93.25$\pm$0.08 & 93.37$\pm$0.16 & 93.31$\pm$0.04 \\

        \cmidrule(lr){1-1}
        \cmidrule(lr){2-2}
        \cmidrule(lr){3-3}
      \cmidrule(lr){4-12}

         \multirow{5}{*}{CIFAR-10}  & \multirow{5}{*}{ResNet-18} & Lottery Pools (interpolation) & {94.95$\pm$0.22} & \textbf{95.09$\pm$0.11} & {94.97$\pm$0.10} & {94.90$\pm$0.22} & {94.94$\pm$0.22} & \textbf{95.16$\pm$0.07} & \textbf{95.07$\pm$0.11} & \textbf{95.05$\pm$0.08} & \textbf{95.16$\pm$0.10} \\ 
         
          & & Lottery Pools (Average)   &  \textbf{95.06$\pm$0.07} & 95.02$\pm$0.11 & \textbf{95.04$\pm$0.17} & \textbf{95.10$\pm$0.06} & \textbf{95.07$\pm$0.10} & 95.09$\pm$0.07 & 95.07$\pm$0.10& 95.03$\pm$0.06 & 94.93$\pm$0.20\\ 
          
          & & SWA   &  94.97$\pm$0.07 & 94.97$\pm$0.07 & 94.96$\pm$0.07 & 94.97$\pm$0.06 & 94.98$\pm$0.06 & 95.05$\pm$0.06 & 95.05$\pm$0.06 & 95.03$\pm$0.10& 95.05$\pm$0.10 \\ 
          
          & & EMA (0.95)  & 94.61$\pm$0.17 & 94.76$\pm$0.09 & 94.76$\pm$0.11 & 94.88$\pm$0.12 & 94.92$\pm$0.04 & 94.98$\pm$0.07 & 95.00$\pm$0.13 & 95.00$\pm$0.09 & 95.05$\pm$0.08  \\ 
          
        & & Original LTs   & 94.23$\pm$0.21 & 94.04$\pm$0.05 & 94.26$\pm$0.21 & 94.44$\pm$0.12 & 94.33$\pm$0.22 & 94.39$\pm$0.27 & 94.48$\pm$0.03 & 94.52$\pm$0.12 & 94.43$\pm$0.00 \\

        \bottomrule
    \end{tabular}}

    \label{tab:performance_deberta_1}

\end{table*}

\begin{table*}[h]
    \centering
    \caption{Accuracy (\%) of Lottery Pools,  the original Lottery Tickets, EMA and SWA on CIFAR-10/100  (2).}

    \label{tab:cifar_results_2}
    \resizebox{.8\textwidth}{!}{
    \begin{tabular}{l|c|c| ccccc ccccc}
        \hline
        \multirow{2}{*}{Dataset}  &  \multirow{2}{*}{Network} & \multirow{2}{*}{Method}  & \multicolumn{10}{c}{Weights Remaining\%} \\
         \cmidrule(lr){4-13}
        & &   &   13.42& 10.74&  8.59&  6.87&  5.50&  4.40&  3.52& 2.81&  2.25&  1.80 \\
        \cmidrule(lr){1-1}
        \cmidrule(lr){2-2}
        \cmidrule(lr){3-3}
        \cmidrule(lr){4-13}
        
          \multirow{5}{*}{CIFAR-100}  & \multirow{5}{*}{VGG-16} & Lottery Pools (interpolation) & {73.21$\pm$0.10} & \textbf{73.14$\pm$0.26} & \textbf{73.27$\pm$0.20} & \textbf{73.16$\pm$0.23} & \textbf{72.79$\pm$0.16} & \textbf{72.62$\pm$0.27} & \textbf{72.38$\pm$0.59} & \textbf{72.20$\pm$0.31} & \textbf{71.41$\pm$0.42} & \textbf{70.89$\pm$0.37} \\ 
         
          & & Lottery Pools (Average)   &   73.04$\pm$0.18 & 72.81$\pm$0.06 & 73.01$\pm$0.30& 72.85$\pm$0.22 & 72.75$\pm$0.11 & 72.47$\pm$0.22 & 72.06$\pm$0.40& 72.10$\pm$0.52 & 71.35$\pm$0.40& 70.08$\pm$0.06\\ 
          
          & & SWA   &  \textbf{73.35$\pm$0.25} & 73.12$\pm$0.28 & 72.92$\pm$0.24 & 72.39$\pm$0.13 & 71.81$\pm$0.14 & 71.02$\pm$0.28 & 69.99$\pm$0.22 & 68.57$\pm$0.36 & 66.41$\pm$0.25 & 62.95$\pm$0.17 \\ 
          
          & & EMA (0.95)  &  73.28$\pm$0.15 & 73.20$\pm$0.19 & 72.95$\pm$0.36 & 72.66$\pm$0.14 & 72.12$\pm$0.23 & 71.76$\pm$0.18 & 71.49$\pm$0.23 & 70.54$\pm$0.14 & 69.57$\pm$0.17 & 67.38$\pm$0.27\\ 
          
        & & Original LTs   & 71.68$\pm$0.34 & 72.03$\pm$0.32 & 71.73$\pm$0.26 & 71.83$\pm$0.08 & 71.87$\pm$0.24 & 71.40$\pm$0.16 & 71.57$\pm$0.30& 71.24$\pm$0.31 & 71.14$\pm$0.25 & 70.08$\pm$0.06 \\ 

        \cmidrule(lr){1-1}
        \cmidrule(lr){2-2}
        \cmidrule(lr){3-3}
        \cmidrule(lr){4-13}

          \multirow{5}{*}{CIFAR-100}  & \multirow{5}{*}{ResNet-18} & Lottery Pools (interpolation)  &   \textbf{75.58$\pm$0.24} & \textbf{75.36$\pm$0.10} & {75.19$\pm$0.11} & \textbf{74.91$\pm$0.22} & \textbf{74.74$\pm$0.10} & \textbf{74.03$\pm$0.22} & \textbf{73.73$\pm$0.15} & \textbf{73.70$\pm$0.34} & \textbf{73.11$\pm$0.07} & \textbf{72.57$\pm$0.17} \\ 
         
          & & Lottery Pools (Average)   & 75.13$\pm$0.46 & 75.11$\pm$0.09 & 75.05$\pm$0.24 & 74.59$\pm$0.25 & 74.33$\pm$0.24 & 73.98$\pm$0.19 & 73.72$\pm$0.09 & 73.23$\pm$0.17 & 73.03$\pm$0.18 & 72.54$\pm$0.05 \\ 
          
          & & SWA   &  75.49$\pm$0.33 & 75.25$\pm$0.22 & 75.06$\pm$0.15 & 74.48$\pm$0.07 & 73.86$\pm$0.18 & 73.20$\pm$0.06 & 72.07$\pm$0.05 & 70.49$\pm$0.11 & 68.36$\pm$0.18 & 64.39$\pm$0.32 \\ 
          
          & & EMA (0.95)  & 75.52$\pm$0.26 & 75.41$\pm$0.09 & \textbf{75.22$\pm$0.04} & 74.81$\pm$0.07 & 74.25$\pm$0.13 & 73.88$\pm$0.19 & 73.42$\pm$0.12 & 72.55$\pm$0.21 & 71.55$\pm$0.12 & 70.12$\pm$0.29 \\ 
          
        & & Original LTs   & 74.40$\pm$0.08 & 74.28$\pm$0.20& 74.14$\pm$0.06 & 73.93$\pm$0.16 & 73.80$\pm$0.12 & 73.69$\pm$0.05 & 73.35$\pm$0.27 & 73.00$\pm$0.44 & 72.80$\pm$0.17 & 72.30$\pm$0.12\\ 

        \cmidrule(lr){1-1}
        \cmidrule(lr){2-2}
        \cmidrule(lr){3-3}
        \cmidrule(lr){4-13}
        
          \multirow{5}{*}{CIFAR-10}  & \multirow{5}{*}{VGG-16} & Lottery Pools (interpolation)   &{94.06$\pm$0.06} & \textbf{94.06$\pm$0.05} & \textbf{94.09$\pm$0.11} & \textbf{94.06$\pm$0.14} & \textbf{94.05$\pm$0.14} & \textbf{93.98$\pm$0.07} & \textbf{93.97$\pm$0.09} & {93.89$\pm$0.10} & \textbf{93.88$\pm$0.09} & \textbf{93.88$\pm$0.09}  \\ 
         
          & & Lottery Pools (Average)   &  93.97$\pm$0.15 & 93.99$\pm$0.08 & 94.05$\pm$0.14 & 94.03$\pm$0.06 & 94.02$\pm$0.11 & 93.85$\pm$0.05 & 93.96$\pm$0.17 & \textbf{93.94$\pm$0.07} & 93.78$\pm$0.16 & 93.77$\pm$0.18\\ 
          
          & & SWA   &  \textbf{94.11$\pm$0.07} & 94.03$\pm$0.05 & 94.02$\pm$0.04 & 93.97$\pm$0.08 & 93.86$\pm$0.08 & 93.72$\pm$0.08 & 93.53$\pm$0.20& 93.20$\pm$0.15 & 92.62$\pm$0.06 & 92.07$\pm$0.06 \\ 
          
          & & EMA (0.95)  &  93.95$\pm$0.06 & 93.99$\pm$0.07 & 94.06$\pm$0.16 & 93.90$\pm$0.02 & 93.94$\pm$0.06 & 93.80$\pm$0.18 & 93.64$\pm$0.07 & 93.46$\pm$0.14 & 93.32$\pm$0.15 & 93.05$\pm$0.13 \\ 
          
        & & Original LTs   &93.26$\pm$0.05 & 93.13$\pm$0.07 & 93.54$\pm$0.04 & 93.31$\pm$0.13 & 93.37$\pm$0.15 & 93.41$\pm$0.02 & 93.44$\pm$0.19 & 93.38$\pm$0.02 & 93.38$\pm$0.09 & 93.28$\pm$0.04 \\ 

        \cmidrule(lr){1-1}
        \cmidrule(lr){2-2}
        \cmidrule(lr){3-3}
        \cmidrule(lr){4-13}
        
          \multirow{5}{*}{CIFAR-10}  & \multirow{5}{*}{ResNet-18} & Lottery Pools(interpolation)& {95.01$\pm$0.12} & \textbf{95.10$\pm$0.14} & \textbf{95.09$\pm$0.14} & \textbf{95.08$\pm$0.04} & \textbf{94.97$\pm$0.11} & \textbf{94.86$\pm$0.09} & \textbf{94.81$\pm$0.10} & \textbf{94.70$\pm$0.12} & \textbf{94.63$\pm$0.14} & {94.39$\pm$0.04} \\ 
         
          & & Lottery Pools (Average)   &  94.91$\pm$0.14 & 94.91$\pm$0.12 & 95.01$\pm$0.11 & 94.82$\pm$0.13 & 94.93$\pm$0.08 & 94.80$\pm$0.21 & 94.71$\pm$0.14 & 94.68$\pm$0.10& 94.51$\pm$0.14 & \textbf{94.43$\pm$0.08}\\ 
          
          & & SWA   &  94.98$\pm$0.10& 95.03$\pm$0.11 & 95.08$\pm$0.10& 94.95$\pm$0.10& 94.87$\pm$0.07 & 94.79$\pm$0.08 & 94.59$\pm$0.18 & 94.28$\pm$0.14 & 93.96$\pm$0.20& 93.28$\pm$0.53  \\ 
          
          & & EMA (0.95)  & \textbf{95.06$\pm$0.15} & 95.09$\pm$0.08 & 94.97$\pm$0.13 & 95.01$\pm$0.04 & 94.91$\pm$0.08 & 94.83$\pm$0.09 & 94.73$\pm$0.20& 94.58$\pm$0.11 & 94.34$\pm$0.18 & 93.99$\pm$0.24 \\ 
          
        & & Original LTs   &  94.54$\pm$0.23 & 94.52$\pm$0.12 & 94.50$\pm$0.13 & 94.53$\pm$0.08 & 94.49$\pm$0.14 & 94.38$\pm$0.15 & 94.31$\pm$0.13 & 94.32$\pm$0.15 & 94.13$\pm$0.25 & 94.11$\pm$0.23\\

        \bottomrule
    \end{tabular}}

    \label{tab:performance_deberta_2}

\end{table*}

\begin{table*}[h]
    \centering
    \caption{Accuracy (\%) of Lottery Pools,  the original Lottery Tickets, EMA and SWA on ImageNet.}

    \label{tab:imgnet_result}
    \resizebox{.65\textwidth}{!}{
    \begin{tabular}{l|c|c| ccccc cccc}
        \hline
        \multirow{2}{*}{Dataset}  &  \multirow{2}{*}{Network} & \multirow{2}{*}{Method}  & \multicolumn{9}{c}{Weights Remaining\%} \\
         \cmidrule(lr){4-12}
        & &   &   100& 80& 64& 51.2 & 40.96& 32.77& 26.21& 20.97& 16.78   \\
        \cmidrule(lr){1-1}
        \cmidrule(lr){2-2}
        \cmidrule(lr){3-3}
        \cmidrule(lr){4-12}
        
          \multirow{5}{*}{ImageNet}  & \multirow{5}{*}{ResNet-18} & Lottery Pools (interpolation)  & \textbf{70.68} & \textbf{70.68} & \textbf{70.70}  & \textbf{70.73} & \textbf{70.64} & \textbf{70.57} & \textbf{70.34} & \textbf{70.04} & \textbf{69.68} \\ 
         
          & & Lottery Pools (Average)   &69.96 & 70.13 & 70.62 & 70.63 & 70.51 & 70.53 & 70.20  & 69.97 & 69.63  \\ 
            
          & & SWA   &  69.60 &  69.60 &  69.60 &  69.48 &  69.34 &  69.09 &  68.36 &  67.69 &  66.46  \\ 
          
          & & EMA (0.95)   & 68.36 &  68.36 &  68.35 &  68.36 &  68.35 &  68.36 &  59.74 &  52.56 &  43.99  \\  
          
        & & Original LTs   & 69.96 & 70.13 & 70.33 & 70.42 & 70.47 & 70.48 & 70.20  & 69.97 & 69.63\\ 

        \cmidrule(lr){1-1}
        \cmidrule(lr){2-2}
        \cmidrule(lr){3-3}
      \cmidrule(lr){4-12}

            \multirow{5}{*}{ImageNet}  & \multirow{5}{*}{ResNet-34} & Lottery Pools (interpolation)  & \textbf{74.36} & \textbf{74.36} & \textbf{74.39} & \textbf{74.32} & \textbf{74.26} & \textbf{74.23} & \textbf{74.15} & \textbf{73.99} & \textbf{73.58}  \\ 
         
          & & Lottery Pools (Average)   &  {73.61} & {74.17} & {74.19} & {74.20}  & {74.22} & {74.17} & {74.01} & {73.95} & {73.53}  \\ 
    
          & & SWA   &   73.56 &  73.75 &  73.76 &  73.80 &  73.75 &  73.51 &  73.25 &  72.45 &  71.41 \\ 
          
          & & EMA (0.95)   &  72.50 &  72.51 &  72.50 &  72.50 &  72.50 &  72.53 &  63.91 &  57.83 &  49.33 \\ 

        & & Original LTs   & 73.61 & 73.73 & 73.86 & 74.01 & 74.05 & 73.91 & 74.01 & 73.95 & 73.53  \\

        \bottomrule
    \end{tabular}}

    \label{tab:performance_deberta}

\end{table*}

\clearpage
\newpage

\end{document}